\pdfoutput=1
\documentclass[11pt]{article}

\usepackage[final]{acl}

\usepackage{times}
\usepackage{latexsym}

\usepackage[T1]{fontenc}
\usepackage[utf8]{inputenc}

\usepackage{microtype}

\usepackage{inconsolata}

\usepackage{graphicx}

\usepackage{amsmath}
\usepackage{amsthm}

\usepackage{amssymb}
\usepackage{amsfonts}
\usepackage{bm}
\usepackage{caption}
\usepackage{multirow}
\usepackage[english]{babel}
\usepackage{adjustbox}
\usepackage{subfigure}
\usepackage{booktabs}
\usepackage{enumitem}
\usepackage{array}
\usepackage{algorithm}
\usepackage{listings}
\usepackage{algpseudocode}

\usepackage{titling}

\usepackage{marvosym}

\usepackage[frozencache,cachedir=.]{minted}
\usepackage{mdframed}
\usepackage{tcolorbox}

\tcbuselibrary{minted,skins, breakable}

\setlist[itemize]{topsep=0pt, partopsep=0pt, parsep=0pt, itemsep=0pt}

\usepackage{xcolor}
\definecolor{mylightblue}{RGB}{100,149,237} % 定义一个自定义的浅蓝色，这里使用的是LightBlue的RGB代码

\tcbset{
  enhanced, % 使用增强皮肤
  colback=white!200!black, % 背景颜色
  colframe=mylightblue, % 边框颜色
  colbacktitle=mylightblue, % 标题背景颜色
  title filled, % 填充标题背景
  coltitle=white, % 标题文字颜色
  fonttitle=\bfseries, % 标题字体
  arc=3mm, % 边框圆角
  outer arc=3mm, % 外边框圆角
  boxrule=0.5mm, % 边框宽度
  toprule=0.5mm, % 顶部边框宽度
  bottomrule=0.5mm, % 底部边框宽度
  titlerule=0.5mm, % 标题下方边框宽度
  drop fuzzy shadow, % 阴影效果
  minted options={ % minted 设置
    breaklines, % 自动换行
    fontsize=\footnotesize, % 字体大小
    linenos, % 显示行号
    numbersep=2mm, % 行号间隔
    framesep=1.5mm, % 内容与框架的间隔
  },
}

\theoremstyle{definition}

\usepackage{color}

\usepackage{colortbl}

\newcount\Comments  % 1 suppresses notes to selves in text
\usepackage{color}
\definecolor{purple}{rgb}{1,0,1}

\AtBeginDocument{%
  \providecommand\BibTeX{{%
    \normalfont B\kern-0.5em{\scshape i\kern-0.25em b}\kern-0.8em\TeX}}}

\makeatletter
\def\blfootnote{\xdef\@thefnmark{}\@footnotetext}
\makeatother

\title{\raisebox{-0.7em}{\includegraphics[width=0.08\textwidth]{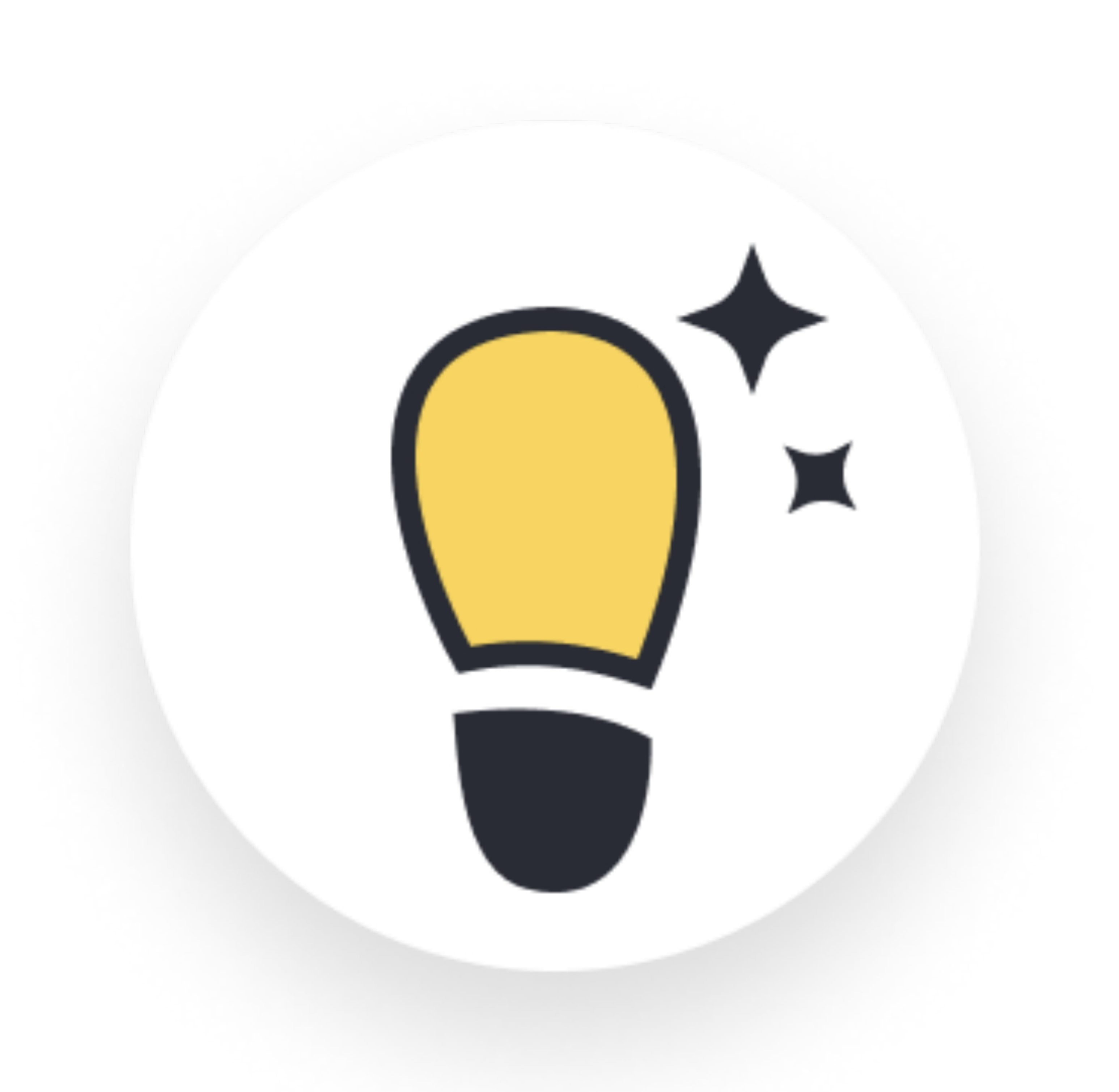}}\textsc{ItiNera}: Integrating Spatial Optimization with Large Language Models for Open-domain Urban Itinerary Planning}

\author{
\textbf{Yihong Tang}\textsuperscript{1,2$*$}, \
\textbf{Zhaokai Wang}\textsuperscript{1,3$*$}, \
\textbf{Ao Qu}\textsuperscript{1,4$\heartsuit*$}, \
\textbf{Yihao Yan}\textsuperscript{1$*$}\textbf{,}  \
\textbf{Zhaofeng Wu}\textsuperscript{1,4} \\
\textbf{Dingyi Zhuang}\textsuperscript{1,4}\textbf{,} \
\textbf{Jushi Kai}\textsuperscript{3}\textbf{,} \
\textbf{Kebing Hou}\textsuperscript{1}\textbf{,}  \
\textbf{Xiaotong Guo}\textsuperscript{1,4}\textbf{,} \
\textbf{Han Zheng}\textsuperscript{1,4} \\
\textbf{Tiange Luo}\textsuperscript{1,5}\textbf{,}\
\textbf{Jinhua Zhao}\textsuperscript{4\Letter}\textbf{,}\
\textbf{Zhan Zhao}\textsuperscript{2\Letter}\textbf{,} 
\textbf{Wei Ma}\textsuperscript{6\Letter} \
\\
[2mm]
\normalsize \textsuperscript{1}Tutu AI \quad
\textsuperscript{2}University of Hong Kong \quad
\textsuperscript{3}Shanghai Jiao Tong University 
\\
\normalsize 
\textsuperscript{4}Massachusetts Institute of Technology
\quad
\textsuperscript{5}University of Michigan 
\\
\normalsize 
\textsuperscript{6}The Hong Kong Polytechnic University
\\
[1.mm]
\tt\small yihongt@connect.hku.hk \quad
wangzhaokai@sjtu.edu.cn \quad
qua@mit.edu \quad yanyihao@tutu-ai.com \\
\tt\small 
jinhua@mit.edu \quad zhanzhao@hku.hk \quad wei.w.ma@polyu.edu.hk
}

\begin{document}

\maketitle

\blfootnote{\vspace{-2.2mm}\\ \noindent$^{*}$Equal contribution. \textsuperscript{\Letter}Corresponding authors. \textsuperscript{$\heartsuit$}Project lead.}
% \blfootnote{Paper in Proceedings of EMNLP 2024.}

\everymath{\small} % 对行内公式应用
\everydisplay{\small} % 对行间公式应用

\begin{abstract}

Citywalk, a recently popular form of urban travel, requires genuine personalization and understanding of fine-grained requests compared to traditional itinerary planning.
In this paper, we introduce the novel task of Open-domain Urban Itinerary Planning (OUIP), which generates \emph{personalized} urban itineraries from user requests in natural language.
We then present \textsc{ItiNera}, an OUIP system that integrates spatial optimization with large language models to provide customized urban itineraries based on user needs. This involves decomposing user requests, selecting candidate points of interest (POIs), ordering the POIs based on cluster-aware spatial optimization, and generating the itinerary.
Experiments on real-world datasets and the performance of the deployed system demonstrate our system's capacity to deliver personalized and spatially coherent itineraries compared to current solutions.
Source codes of \textsc{ItiNera} are available at \url{https://github.com/YihongT/ITINERA}.

\end{abstract}

\section{Introduction}

\begin{figure*}[t!]
    \centering
    \includegraphics[width=0.7\linewidth]{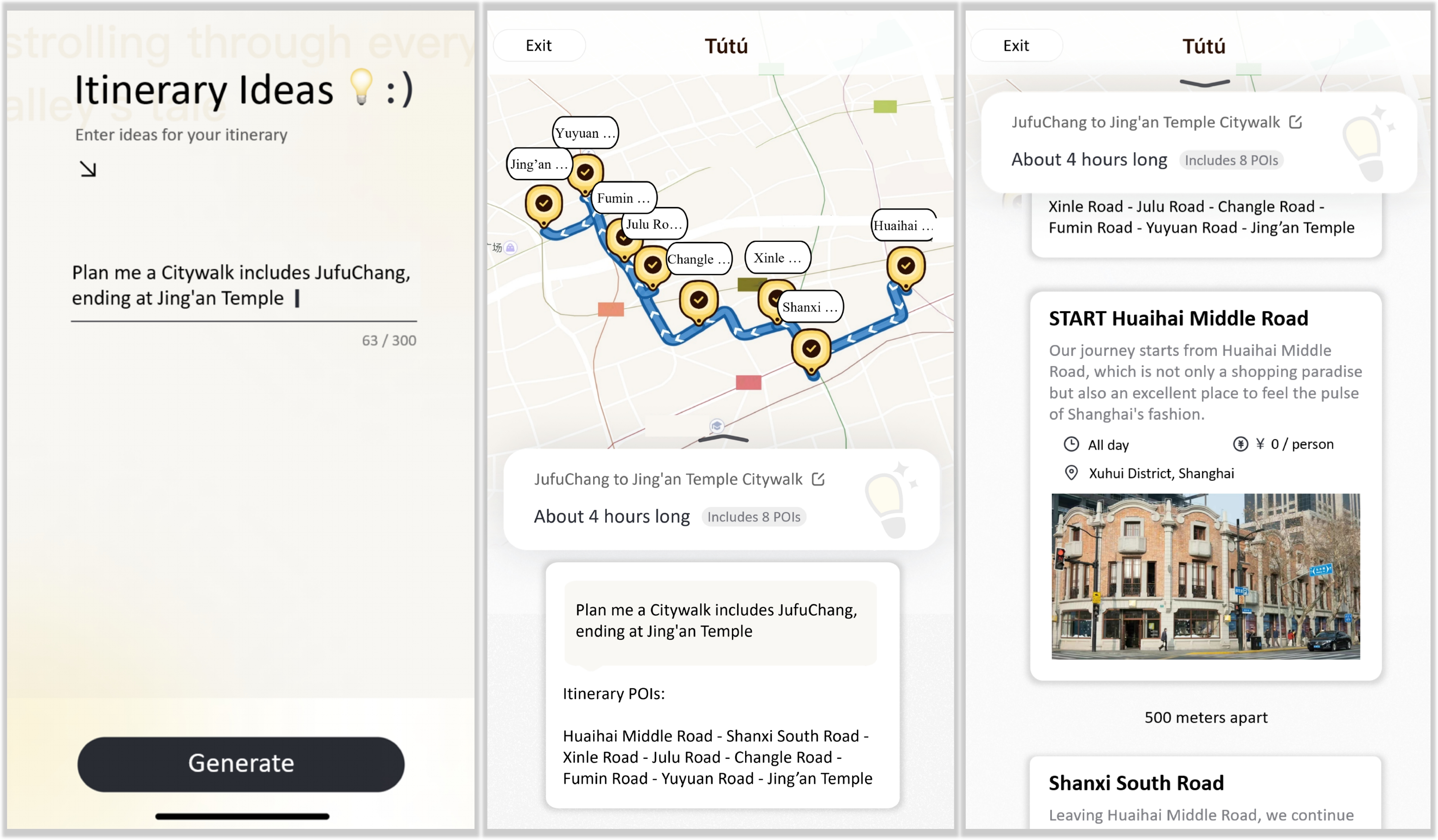}
    {
    \captionsetup{belowskip=-0.5em}
    \caption{The OUIP problem and the OUIP system.}
    \label{fig:itiplan}
    }
    
\end{figure*}

As a novel form of urban travel, citywalk \cite{Germano_2023} invites travelers to wander through city streets and immerse themselves in local culture, offering a more dynamic, immersive, and fine-grained travel experience compared to traditional tourism. Planning a citywalk is a complex urban itinerary planning problem \cite{halder2024survey}, involving travel-related information gathering, POI selection, route mapping, and customization for diverse user needs. 
Specifically, citywalk differs from traditional tourism by (1) Dynamic Information: involving rapidly changing POIs and needing up-to-date information on temporary events, (2) Personalization: prioritizing individual preferences over widely recognized POIs, and (3) Diverse Constraints: considering complex constraints like personal interests and accessibility requirements. An example of the OUIP problem is shown in Fig.~\ref{fig:itiplan}.

Existing itinerary planning studies focus on traditional tourism. They consider coarse-grained user requirements such as geographical constraints \cite{rani2018development} and time budgets \cite{hsueh2019personalized} to improve the quality of an itinerary \cite{chen2013automatic,sylejmani2017planning}. While these optimization-based approaches maintain the quality of POIs and spatial coherency, they struggle to address dynamic and detailed personal demands, leading to itineraries that lack personalization and diversity.

Recently, large language models (LLMs)~\cite{openai2023gpt} have shown impressive applications in understanding user needs and following instructions. However, their limitations in itinerary planning are evident \cite{Xie2024TravelPlanner}: (1) Pure LLMs cannot refer to specific POI lists, resulting in outdated or hallucinated POIs. (2) LLMs lack the optimization capabilities required for planning tasks, leading to suboptimal itineraries. Consequently, LLM-generated itineraries can be circuitous, lack detail, and include impractical information.

To address these limitations, in this work, we first define the Open-domain Urban Itinerary Planning (OUIP) problem, which involves \emph{generating personalized travel itineraries based on user requests in natural language}. Then, we propose \textsc{ItiNera}, a holistic OUIP system that integrates spatial optimization with LLMs. 
\textsc{ItiNera} comprises five LLM-assisted modules: \textit{User-owned POI Database Construction} (UPC), \textit{Request Decomposition} (RD), \textit{Preference-aware POI Retrieval} (PPR), \textit{Cluster-aware Spatial Optimization} (CSO), and \textit{Itinerary Generation} (IG), to deliver personalized and spatially coherent itineraries.

Our overall contributions are:
\begin{itemize}[leftmargin=*]
\item We introduce the OUIP problem to provide personalized urban travel itineraries based on users' natural language inputs and propose metrics to measure the quality of generated itineraries.
\item We develop \textsc{ItiNera}, an LLM-based OUIP system that combines spatial optimization with LLMs to create fine-grained urban itineraries tailored to users' requests.
\item Extensive experiments on the real-world dataset and performance of the deployed system show that \textsc{ItiNera} creates personalized, spatially coherent, and high-quality urban itineraries that meet user requirements.
\end{itemize}

\begin{figure*}[t]
    \centering
    \includegraphics[width=\textwidth]{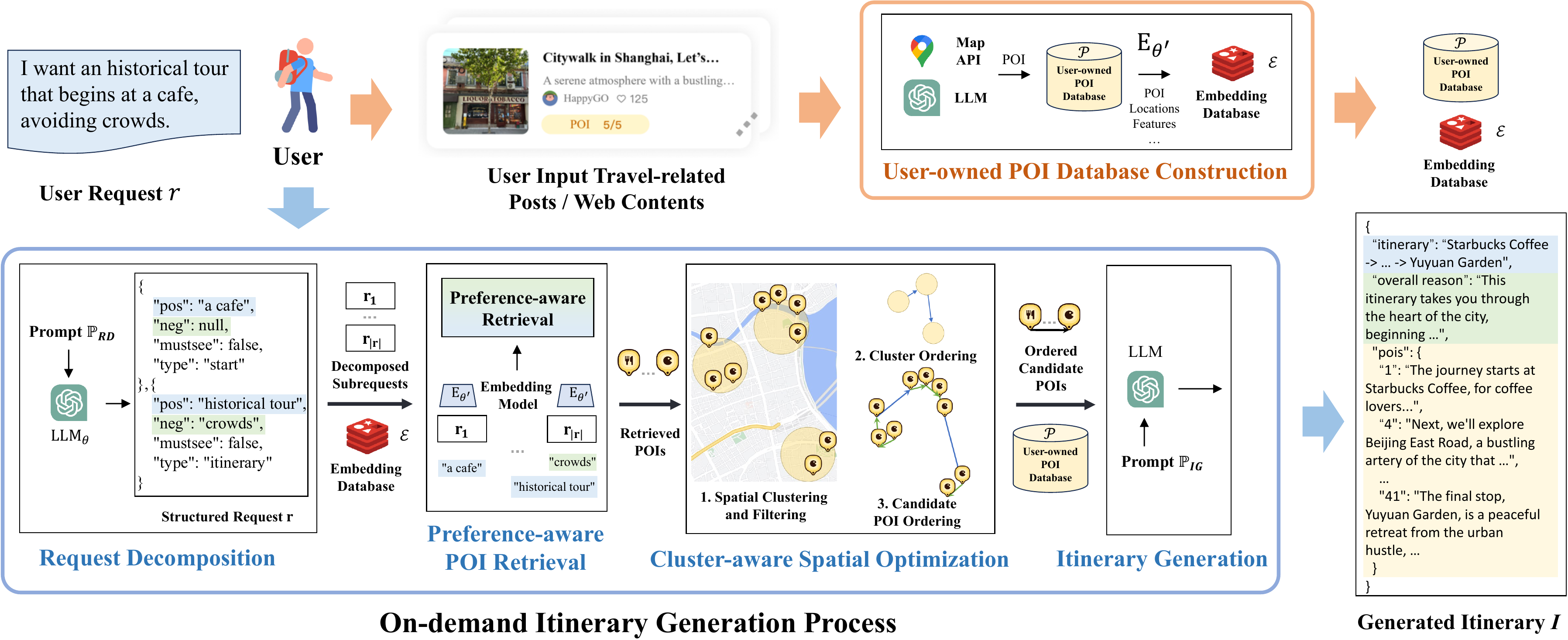}
    \vspace{-0.3em}
    {
    \caption{An overview of the proposed \textsc{ItiNera} system.} 
    \label{fig:architecture}
    }
    \vspace{-0.5em}
\end{figure*}

\vspace{-1mm}
\section{Related Work}

\vspace{-1mm}
\paragraph{LLMs in Urban Applications}

Since ChatGPT, LLMs have demonstrated strong knowledge and reasoning capabilities. 
Recent studies highlight the potential of LLMs in urban data processing \cite{yan2023urban} and urban planning \cite{zhou2024large}. 
These works reveal LLMs' capabilities in predicting human mobility patterns \cite{mo2023large,xue2022leveraging} and emphasize their predictive strength \cite{wang2023would}. 
In transportation, LLMs contribute to traffic safety analysis \cite{zheng2023trafficsafetygpt}, enhance traffic forecasting \cite{de2023llm}, and automate accident report generation \cite{zheng2023chat}
, showing their applicability in urban transportation.
Leveraging LLMs for travel planning has recently gained public interest. 
TravelPlanner \cite{Xie2024TravelPlanner} proposes a sandbox environment with various tools for benchmarking LLMs on multi-day travel planning, revealing LLMs' current limitations for complex planning tasks. 
Unlike TravelPlanner, our system focuses on fine-grained OUIP, addressing urban itinerary planning within a single day, but can be seamlessly extended to multi-day travel planning.

\paragraph{Itinerary Planning (IP)}

Current research on IP focuses on creating itineraries based on a set of POIs. Some methods directly optimize the spatial utilities of the itinerary, while others define IP as an Orienteering Problem (OP) and consider constraints that include time \cite{zhang2018itinerary,hsueh2019personalized}, space \cite{rani2018development}, must-see POIs \cite{taylor2018travel}, categories \cite{bolzoni2014efficient}
, and their combinations \cite{gionis2014customized,yochum2020adaptive}, 
to indirectly ensure the spatial coherence and quality of the itinerary. However, their ability to personalize is limited.
Recommendation-based methods \cite{ho2022poibert,tang2022hgarn} could be applied to the IP task, but they depend on historical user behavior data. 
Overall, existing IP methods struggle with open-domain, user natural-language inputs, failing to generate personalized itineraries, making them unsuitable for OUIP.

\vspace{-2mm}
\section{Methodology}
\vspace{-1mm}

We formalize the OUIP problem and explain how \textsc{ItiNera} generates itineraries, as shown in Fig. \ref{fig:architecture}.

\subsection{Open-domain Urban Itinerary Planning (OUIP) Problem}

To enable personalized OUIP, an open-domain system is essential. Such a system allows users to freely express their diverse requirements and expectations, enabling the planning of urban itineraries tailored to their specific needs and purposes.

Formally, given a user request $r$ in natural language and the user-owned POI database $\mathcal{P}=\{p_j\}_{j=1}^{N}$, the OUIP problem aims to find an itinerary generator $\mathcal{G}$ to select and order a subset of POIs from $\mathcal{P}$ to create a coherent travel itinerary $I$ as an ordered list of POIs that optimally aligns with the user's requests $r$ while adhering to spatio-temporal considerations: $I \sim \mathcal{G}\left(\mathcal{P} | r\right)$.

\vspace{-2mm}
\subsection{User-owned POI Database Construction}
\label{sec:ppa}

Travelers often have specific places they want to visit or particular requirements for the POIs in their itinerary. To ensure a fully personalized itinerary planning process, we have developed an automated pipeline that extracts POIs and relevant details from social media, catering to these individual needs.
Users can input travel post links, and the pipeline uses LLMs to extract POIs and their descriptions, calls Map APIs and embedding models to obtain their locations and embeddings, and integrates the information into the user-owned POI database $\mathcal{P}$ and embedding database $\mathcal{E}$. 

The user-owned POIs enable users to create personalized POI databases, maintain timely POI information, and customize travel itineraries, enhancing itinerary experiences.
We execute a daily routine to aggregate POIs from trending posts across multiple cities and maintain an up-to-date, dynamic and comprehensive POI database. 
This database serves as the initial set of POIs for any new user, substantially mitigating the potential cold start issue for POI acquisition.
The pipeline and the database format are detailed in \S\ref{supple:pipe} and \S\ref{supple:dataset}.

\vspace{-1mm}
\subsection{Request Decomposition}
\vspace{-1mm}

Upon receiving user requests, we use LLMs to structure and extract information. A single user request $r$ can be decomposed into multiple independent subrequests, each reflecting preferences at different levels and classified by granularity, specificity, and attitude.
Granularity includes (1) POI-level and (2) itinerary-level subrequests. Specificity has (1) specific and (2) vague subrequests. Attitude distinguishes (1) positive subrequests (likes) and (2) negative subrequests (dislikes).

We prompt the LLM to decompose the user request $r$ based on these categories. Formally, we obtain the resulting set of structured subrequests $\mathcal{R}=\{\mathbf{r}_i\}_{i=1}^{|\mathcal{R}|}$ through: $\mathcal{R} \sim \operatorname{LLM}\left( \mathbb{P}_{RD}(r)\right)$, where $\mathbb{P}_{RD}$ wraps the request $r$ with instructions and examples (see \S\ref{supple:prompt_decom}).
Here, `pos' and `neg' indicate attitude-level subrequests. `Mustsee' is a boolean for specificity-level subrequests, and `type' indicates granularity-level subrequests, which can be one of ["start" (POI-level origin), "end" (POI-level destination), "POI" (POI-level), "itinerary" (itinerary-level)].

\vspace{-1mm}
\subsection{Preference-aware POI Retrieval}
\vspace{-1mm}

After obtaining decomposed subrequests, we select POIs from the user-owned POI database $\mathcal{P}$ that match their preferences. While LLMs excel in language comprehension, they are limited by context window size and input token cost. Given the vast amount of POI data and LLM inference speed limitations, we design a preference-aware embedding-based retrieval approach.
For a subrequest $\mathbf{r}_i$, we first use an embedding model $\operatorname{E}_{\theta^\prime}$ to encode the `pos' and `neg' fields: $\mathbf{e}^{\operatorname{pos}}_{i}=\operatorname{E}_{\theta^\prime}\left(\mathbf{r}_{i}^{\operatorname{pos}}\right); \, \mathbf{e}^{\operatorname{neg}}_{i}=\operatorname{E}_{\theta^\prime}\left(\mathbf{r}_{i}^{\operatorname{neg}}\right)$, 
where $\theta^\prime$ denotes the parameters of the $\operatorname{E}$, and $\mathbf{e}^{\operatorname{neg}}_{i}, \mathbf{e}^{\operatorname{pos}}_{i} \in \mathbb{R}^{d}$ are embeddings.

Ideally, we want the queried POIs to best fit the positive subrequest while avoiding the negative subrequest. To achieve this, we use the positive embedding $\mathbf{e}^{\operatorname{pos}}_{i}$ to retrieve $k$ POIs from $\mathcal{P}$ to obtain $\mathcal{P}^{\operatorname{pos}}_{i}=\{p^{\operatorname{pos}}_{i, j}\}_{j=1}^{k}$ and the corresponding embeddings $\mathcal{E}_{i}^{\operatorname{pos}}=\{\mathbf{e}^{\operatorname{pos}}_{i, j}\}_{j=1}^{k}$ from $\mathcal{E}$ with top similarity scores $\mathcal{S}^{\operatorname{pos}}_{i}=\{s^{\operatorname{pos}}_{i, j}\}_{j=1}^{k}$, where $p^{\operatorname{pos}}_{i, j}$ and $s^{\operatorname{pos}}_{i, j}$ represent the $j$th POI and score for $i$th positive sub-request.
Next, we compute the similarity scores between $\mathcal{E}^{\operatorname{pos}}_i$ and $\mathbf{e}^{\operatorname{neg}}_{i}$ and rerank the POIs based on the gap between positive and negative similarity scores. Using $\mathcal{E} \in \mathbb{R}^{N \times d}$ to denote the pre-computed POI embeddings in the user-owned database, the process is:
\setlength{\abovedisplayskip}{0.8pt} 
\setlength{\belowdisplayskip}{0.8pt}
\begin{eqnarray}
\mathcal{P}^{\operatorname{pos}}_{i}, \mathcal{S}^{\operatorname{pos}}_{i}, \mathcal{E}^{\operatorname{pos}}_{i} &=& \operatorname{score}^{k}\left( \mathbf{e}^{\operatorname{pos}}_{i}, \mathcal{E}\right)\\
\mathcal{P}^{\operatorname{neg}}_{i}, \mathcal{S}^{\operatorname{neg}}_{i}, \mathcal{E}^{\operatorname{neg}}_{i} &=& \operatorname{score}\left( \mathbf{e}^{\operatorname{neg}}_{i}, \mathcal{E}^{\operatorname{pos}}_i \right)\\
\mathcal{P}_i, \mathcal{S}_i &=& \operatorname{rank}\left(
\mathcal{P}^{\operatorname{pos}}_{i},  \mathcal{S}^{\operatorname{pos}}_{i}- \mathcal{S}^{\operatorname{neg}}_{i} \right),
\end{eqnarray}
where the $\operatorname{score}(\cdot)$ function measures embedding similarities, and the superscript $k$ indicates it returns the top-$k$ results
. The $\operatorname{rank}(\cdot)$ reorders POIs from highest to lowest similarity scores.

Lastly, we select the top-$k$ POIs with the highest summed scores from the union of all retrieved POIs to form the final set $\mathcal{P}^{rt}$ for the user request $r$:
\begin{equation} \label{eq:ret_final}
    \mathcal{P}^{rt}, \mathcal{S}^{rt}=\operatorname{rank}^{k}\left( \cup_{i=1}^{|\mathcal{R}|} \left(\mathcal{P}_i, \mathcal{S}_i\right)\right) \cup \left(\mathcal{P}^{\operatorname{must}}, \mathcal{S}^{\operatorname{must}}\right),
\end{equation}
where $\mathcal{S}^{\operatorname{must}}$ has large values to ensure must-see POIs are included. During the union process, scores for the same POI under different subrequests are summed to obtain the final score.

\begin{table*}[!h]
\resizebox{\linewidth}{!}{
    \small
    \begin{tabular}{@{}lcccc|ccc|cccc|ccc@{}}
    \toprule
     & \multicolumn{7}{c|}{Shanghai} & \multicolumn{7}{c}{Qingdao} \\ \cmidrule(l){2-15} 
    
    Method & \multicolumn{4}{c|}{Rule-based Metrics} & \multicolumn{3}{c|}{LLM-Eval $\uparrow$ (\%)} & \multicolumn{4}{c|}{Rule-based Metrics} & \multicolumn{3}{c}{LLM-Eval $\uparrow$ (\%)} \\ \cmidrule(l){2-15} 
    
     & \begin{tabular}[c]{@{}c@{}}RR $\uparrow$ \\ (\%)\end{tabular} & \begin{tabular}[c]{@{}c@{}}AM $\downarrow$ \\ (meters)\end{tabular} & OL $\downarrow$ & \begin{tabular}[c]{@{}c@{}}FR $\downarrow$ \\ (\%)\end{tabular} & PQ & IQ & Match
    & \begin{tabular}[c]{@{}c@{}}RR $\uparrow$ \\ (\%)\end{tabular} & \begin{tabular}[c]{@{}c@{}}AM $\downarrow$ \\ (meters)\end{tabular} & OL $\downarrow$ & \begin{tabular}[c]{@{}c@{}}FR $\downarrow$ \\ (\%)\end{tabular} & PQ & IQ & Match
     \\ \midrule

    Ground Truth & / & 124.4 & 0.44 & / & 68.9 & 61.5 & 80.9 & / & 356.6 & 0.31 & / & 75.4 & 63.9 & 71.4 \\
    \midrule

    IP & 6.4 & 1573.3 & 2.96 & / & 30.3 & 26.2 & 17.8 & 7.6 & 4134.3 & 2.86 & / & 23.6 & 16.8 & 20.2 \\
    Ernie-Bot 4.0 & 15.7 & 513.5  & 0.91 & 15.2 & 42.1 & 46.5 & 42.5 & 27.2 & 776.2 & 0.78 & 21.4 & 43.4 & 38.2 & 33.3 \\
    GPT-3.5 & 16.6 & 422.4 & 0.83 & 13.5 & 40.4 & 43.1 & 45.4 & 25.5 & 691.5 & 0.55 & 22.0 & 33.4 & 39.0 & 46.6 \\
    GPT-4 & 18.0 & 267.2 & 0.56 & 8.2 & 45.0 & 48.2 & 46.9 & 27.3 & 569.4 & 0.49 & 19.6 & 46.6 & 48.7 & 48.4 \\
    GPT-4 CoT & 18.4 & 258.3 & 0.49 & 7.5 & / & / & / & 30.2 & 542.6 & 0.43 & 17.8 & / & / & / \\
    \textsc{ItiNera} (ours) & \textbf{31.4} & \textbf{86.0} & \textbf{0.42} & / & \textbf{69.8} & \textbf{64.6} & \textbf{72.0} & \textbf{35.4} & \textbf{225.8} & \textbf{0.26} & / & \textbf{71.2} & \textbf{68.2} & \textbf{67.8} \\  
    \toprule
     & \multicolumn{7}{c|}{Beijing} & \multicolumn{7}{c}{Hangzhou} \\ \cmidrule(l){2-15} 
    
    Method & \multicolumn{4}{c|}{Rule-based Metrics} & \multicolumn{3}{c|}{LLM-Eval $\uparrow$ (\%)} & \multicolumn{4}{c|}{Rule-based Metrics} & \multicolumn{3}{c}{LLM-Eval $\uparrow$ (\%)} \\ \cmidrule(l){2-15} 
    
     & \begin{tabular}[c]{@{}c@{}}RR $\uparrow$ \\ (\%)\end{tabular} & \begin{tabular}[c]{@{}c@{}}AM $\downarrow$ \\ (meters)\end{tabular} & OL $\downarrow$ & \begin{tabular}[c]{@{}c@{}}FR $\downarrow$ \\ (\%)\end{tabular} & PQ & IQ & Match
    & \begin{tabular}[c]{@{}c@{}}RR $\uparrow$ \\ (\%)\end{tabular} & \begin{tabular}[c]{@{}c@{}}AM $\downarrow$ \\ (meters)\end{tabular} & OL $\downarrow$ & \begin{tabular}[c]{@{}c@{}}FR $\downarrow$ \\ (\%)\end{tabular} & PQ & IQ & Match
     \\ \midrule

    Ground Truth & / & 218.3 & 0.53 & / & 61.9 & 57.3 & 77.0 & / & 70.9 & 0.34 & / & 47.5 & 53.2 & 66.3 \\
    \midrule

    IP & 3.3 & 3034.2 & 2.26 & / & 27.8 & 18.2 & 20.4 & 1.8 & 1744.4 & 1.52 & / & 34.8 & 31.4 & 22.5 \\
    Ernie-Bot 4.0 & 18.8 & 379.4  & 0.74 & 12.8 & 31.2 & 34.8 & 32.1 & 12.9 & 605.2 & 1.17 & 24.4 & 43.6 & 34.3 & 38.2 \\
    GPT-3.5 & 19.7 & 347.8 & 0.58 & 14.3 & 29.2 & 40.5 & 43.8 & 14.4 & 665.4 & 1.16 & 19.8 & 41.2 & 40.8 & 32.8 \\
    GPT-4 & 20.6 & 342.6 & 0.52 & 11.1 & 45.4 & 43.6 & 45.2 & 14.8 & 746.1 & 1.09 & 23.2 & 46.2 & 39.6 & 39.4 \\
    GPT-4 CoT & 21.0 & 327.7 & 0.54 & 10.2 & / & / & / & 15.5 & 455.0 & 1.09 & 18.6 & / & / & / \\
    \textsc{ItiNera} (ours) & \textbf{28.4} & \textbf{79.2} & \textbf{0.46} & / & \textbf{59.2} & \textbf{67.6} & \textbf{75.2} & \textbf{21.4} & \textbf{30.5} & \textbf{0.12} & / & \textbf{61.6} & \textbf{65.4} & \textbf{68.3} \\
    \bottomrule

    \end{tabular}
}
\vspace{-0.5em}
\caption{Overall results on four datasets. LLM-evaluated metrics are win rates against GPT-4 CoT. }
\vspace{-0.5em}
\label{tab:main-exp}
\end{table*}

\subsection{Cluster-aware Spatial Optimization}
\subsubsection{Spatial Clustering and Filtering}

\begin{algorithm} 
\footnotesize
{
    \caption{\footnotesize Spatial Clustering \& Candidate POI Selection}\label{algo:spatial_cluster_and_candidate_poi}
}
\begin{algorithmic}[1]
\footnotesize
\Statex \textbf{Input:} Retrieved POI set $\mathcal{P}^{rt}$ with scores $\mathcal{S}^{rt}$, Distance threshold $\tau$, Candidate POIs num threshold $N^c$
\Statex \textbf{Output:} Spatial Clusters $\mathcal{C}$, Candidate POIs $\mathcal{P}^c$

\State \textcolor{darkgray}{// \textsc{Spatial Clustering}} 
\State $G \leftarrow (V, E)$ with $V \leftarrow \mathcal{P}^{rt}, E \leftarrow \emptyset$; $\mathcal{C} \leftarrow \emptyset$; $\mathcal{P}^c \leftarrow \emptyset$
\For{$p_a^{rt}, p_b^{rt} \in V$ with $a \neq b$}
    \If{${distance}(p_a^{rt}, p_b^{rt}) < \tau$}
        \State $E \leftarrow E \cup \{(p_a^{rt}, p_b^{rt})\}$
    \EndIf
\EndFor

\While{$V \neq \emptyset$}
    \State $c \leftarrow \text{largest clique in } G$
    \State $\mathcal{C} \leftarrow \mathcal{C} \cup \{c\}$; $V \leftarrow V \setminus c$
\EndWhile

\State \textcolor{darkgray}{// \textsc{Selection of Candidate POIs}} 
\For{each cluster $c \in \mathcal{C}$}
    \State $s^{rt}_c \leftarrow \sum_{p_j \in c} s_{j}^{rt}$
\EndFor

\State Sort $\mathcal{C}$ in descending order of $\mathcal{S}^c=\{s^{rt}_{c}\}_{c=1}^{\mathcal{C}}$
\While{$|\mathcal{P}^c| < N^c$}
    \State $c_{\text{max}} \leftarrow \arg\max_{c \in \mathcal{C}} s_c^{rt}$
    \State $\mathcal{P}^c \leftarrow \mathcal{P}^c \cup c_{\text{max}}$; $\mathcal{C} \leftarrow \mathcal{C} \setminus \{c_{\text{max}}\}$
\EndWhile

\State \Return $\mathcal{C}, \mathcal{P}^c$
\end{algorithmic}
\end{algorithm}

A spatially coherent itinerary enhances the travel experience by allowing travelers to move efficiently between clusters of POIs, reducing transit time and effort \cite{bolzoni2014efficient}. Therefore, spatially filtering and sequencing the retrieved POIs is essential.
To achieve this, we compute spatial clusters of the retrieved POIs and select candidates based on proximity and matching scores, addressing cluster-aware spatial optimization by solving a hierarchical traveling salesman problem \cite{jiang2014hierarchical}, a common and fundamental spatial reasoning task~\cite{tang2024sparkle}. Given the retrieved POIs $\mathcal{P}^{rt}$, we create a spatial proximity graph $G$ using a distance matrix $D$, where each node is a POI and edges connect nodes within a distance threshold $\tau$. A community detection algorithm divides $G$ into clusters. We iteratively select the cluster with the highest summed similarity score until the total number of selected POIs reaches a threshold $N^c$, forming the candidate POIs $\mathcal{P}^c$ for the user request $r$. The process is detailed in Algo.~\ref{algo:spatial_cluster_and_candidate_poi}.

\subsubsection{POI Ordering via Solving Hierarchical TSP}
\vspace{-3mm}

\begin{algorithm} 
\footnotesize
\caption{\footnotesize Hierarchical TSP for POI Ordering}\label{algo:hierarchical_tsp}
\begin{algorithmic}[1]
\footnotesize
\Statex \textbf{Input:} Spatial clusters $\mathcal{C}$, candidate POIs $\mathcal{P}^c$, distance matrix $D$
\Statex \textbf{Output:} Ordered list of candidate POIs $\mathcal{P}^{\text{order}}$

\State \textcolor{darkgray}{// \textsc{POI Ordering}} 
\State $\mathcal{C}^{\text{order}} \leftarrow \text{SolveTSP}(\mathcal{C}, D)$; $\mathcal{P}^{\text{order}} \leftarrow \emptyset$
\For{each cluster $c$ in $\mathcal{C}^{\text{order}}$}
    \State $p_{\text{start}}^c, p_{\text{end}}^c \leftarrow \text{GetClusterEndpoints}(c, \mathcal{C}^{\text{order}}, D)$
    \State $c_{\text{path}} \leftarrow \text{SolveConstraintTSP}(c, p_{\text{start}}^c, p_{\text{end}}^c, D)$
    \State $\mathcal{P}^{\text{order}} \leftarrow \mathcal{P}^{\text{order}} \cup c_{\text{path}}$
\EndFor

\State \textcolor{darkgray}{// \textsc{Start POI Selection and POI Reordering}} 
\State $p_\text{start} \leftarrow \text{Select}(\mathcal{P}^{\text{order}})$
\State $\mathcal{P}^{\text{order}} \leftarrow \text{Reorder}(\mathcal{P}^{\text{order}}, p^\text{order}_\text{start})$ 

\State \Return $\mathcal{P}^{\text{order}}$
\end{algorithmic}
\end{algorithm}
\vspace{-3mm}

After obtaining the spatial clusters $\mathcal{C}$, we optimize the access order of the candidate POIs for a spatially coherent itinerary by determining the access order of each cluster and solving TSPs within each cluster with start and end POI constraints. Start and end points are selected based on the proximity of POIs in adjacent clusters, as shown in Fig. \ref{fig:architecture}. This process, outlined in Algo.~\ref{algo:hierarchical_tsp}, optimizes and ensures coherent traversal among selected POIs using an efficient hierarchical TSP approach. `SolveTSP' and `SolveTSPWithEndpoints' handle standard and constrained TSPs, respectively, while `GetClusterEndpoints' determines the start and end points for each cluster. The starting point, $p_\text{start}$, is identified by `Select' from subrequests $\mathcal{R}$ or by prompting an LLM with $\mathcal{P}^c$ and $r$. Finally, the `Reorder' function arranges the POIs in the original order of precedence starting from $p_\text{start}$. Further details are in \S\ref{supple:spatial}.

\vspace{-1mm}
\subsection{Itinerary Generation}

Selecting a subset from $\mathcal{P}^{\text{order}}$ ensures a spatially coherent itinerary, but a high-quality itinerary must also meet constraints like time availability and practicality. It should, for example, avoid consecutive restaurant visits and schedule activities appropriately, such as bars in the evening or coffee shops in the morning. Traditional optimization algorithms can become overly complex and lack variability \cite{yochum2020adaptive,taylor2018travel}, hindering itinerary diversity. To address this, we leverage the advanced reasoning and planning capabilities of LLMs to generate final itineraries that meet these diverse constraints.

Specifically, the primary objective of this module is to effectively utilize LLM to select an optimal subset from $\mathcal{P}^{\text{order}}$, which closely aligns with user requests while adhering to various constraints. This process can be formally defined as follows:

\vspace{-1em}
\setlength{\abovedisplayskip}{0.01pt} 
\setlength{\belowdisplayskip}{0.01pt}
\begin{equation}
    I \sim \operatorname{LLM}\left(\mathbb{P}_{IG}\left( \mathbf{r}, \, \mathcal{P}^{\text{order}} \,, I_{ex}\right)\right),
\end{equation}
\vspace{-1.5em}

\noindent where $I_{ex}$ indicates extra natural language input that improves the language quality of the generated itinerary.
The prompt $\mathbb{P}_{IG}$ for generating the final itinerary contains the following parts:
(1) User request information;
(2) Candidate POIs with context;
(3) Task description;
(4) Specific constraints;
(5) Language style;
(6) Output format.
The full prompt is provided in \S\ref{supple:prompt_ig}. 

\vspace{-1mm}
\section{Experiments}

\vspace{-1mm}
\subsection{Experimental Setup}

We collect an urban travel itinerary dataset from four Chinese cities in collaboration with a leading travel agency specializing in single-day citywalk. 
Each data sample contains a user request, the corresponding urban itinerary plan, and detailed POI data.
In total, the dataset covers 1233 top-rated urban itineraries and 7578 POIs.
For detailed data format, sample entries, and key preprocessing methodologies employed, refer to \S\ref{supple:dataset}.

We use GPT-4 for final itinerary generation to ensure quality and GPT-3.5 for other interactions to speed up responses. Our system and data are originally in Chinese, and we provide a translated version in this paper. Additional implementation details are in \S\ref{supple:imple}.

\begin{table*}[!h]
\centering
\resizebox{.95\linewidth}{!}{
    \small
    \begin{tabular}{@{}lcccccccc|ccc@{}}
    \toprule
    \multirow{2}{*}{Variants} & \multirow{2}{*}{UPC} & \multirow{2}{*}{RD} & \multirow{2}{*}{PPR} & \multirow{2}{*}{CSO} & \multirow{2}{*}{IG} & \multicolumn{3}{c|}{Rule-based Metrics} & \multicolumn{3}{c}{LLM-Eval $\uparrow$ (\%)} \\ \cmidrule(l){7-12} 
     &  &  &  &  &  & RR $\uparrow$ & AM $\downarrow$  & OL $\downarrow$ & PQ & IQ & Match \\ \midrule
    GPT-4 CoT & $\times$ & $\times$ & $\times$ & $\times$ & \checkmark & 18.4 & 258.3 & 0.49 & / & / & / \\
    GPT-4 CoT + UPC & \checkmark & $\times$ & \checkmark & $\times$ & \checkmark & 34.2 & 240.2 & 0.52 & 65.5 & 61.8 & 70.6 \\
    \textsc{ItiNera} w/o RD & \checkmark & $\times$ & \checkmark & \checkmark & \checkmark & 22.6 & 35.4 & 0.18 & 68.2 & 61.5 & 60.5 \\
    \textsc{ItiNera} w/o PPR & \checkmark & \checkmark & $\times$ & \checkmark & \checkmark & 28.2 & 84.6 & 0.38 & 66.7 & 63.4 & 62.2 \\
    \textsc{ItiNera} w/o CSO & \checkmark & \checkmark & \checkmark & $\times$ & \checkmark & 32.8 & 242.8 & 1.04 & 72.1 & 60.2 & 74.2 \\
    \textsc{ItiNera} w/ GPT-3.5 & \checkmark & \checkmark & \checkmark & \checkmark & \checkmark & 27.6 & 79.4 & 0.56 & 67.3 & 58.8 & 61.4 \\
    \textsc{ItiNera} w/ LLaMA 3.1 8B & \checkmark & \checkmark & \checkmark & \checkmark & \checkmark & 27.8 & 90.6 & 0.45 & 66.9 & 58.6 & 63.5 \\
    \textsc{ItiNera} (full) & \checkmark & \checkmark & \checkmark & \checkmark & \checkmark & 31.4 & 86.0 & 0.42 & 69.8 & 64.6 & 72.0 \\ \bottomrule
    \end{tabular}
}
\vspace{-0.5em}
\caption{Ablation study on Shanghai dataset.}
\vspace{-0.5em}
\label{tab:ablation}
\end{table*}

\vspace{-2mm}
\subsection{Evaluation Metrics}\label{sec:metric}
\vspace{-1mm}

A satisfactory itinerary must be spatially coherent and aligned with the user's needs, so we designed the following evaluation metrics.

\vspace{-1mm}
\paragraph{Rule-based Metrics}

(1) \textbf{Recall Rate (RR)}: the recall rate of POIs in the ground truth itinerary, which evaluates the accuracy of understanding user requests and recommending personalized POIs. (2) \textbf{Average Margin (AM)}: the average difference per POI between the total distance of the generated itinerary and the shortest distance (via TSP). (3) \textbf{Overlaps (OL)}: the number of self-intersection points in the generated itinerary. AM and OL measure spatial optimization for POI visit order, with lower values being better. (4) \textbf{Fail Rate (FR)}: the percentage of POIs from LLM not matched with queried map service POIs, which assesses the LLM's information accuracy, as failed POIs are inaccessible and impact the user experience.

\vspace{-1mm}
\paragraph{LLM-Evaluated Metrics}

The rule-based metrics are intuitive, but some aspects, like POI appeal and alignment with user requests, are hard to quantify. Thus, we propose several LLM-evaluated metrics: (1) \textbf{POI Quality (PQ)}: how interesting and diverse the POIs are; (2) \textbf{Itinerary Quality (IQ)}: the overall quality and coherence of the itinerary; (3) \textbf{Match}: the alignment between the itinerary and the user request. We use GPT-4 to rank two itineraries and compute the win rate, repeated at least 10 times for reliability. Our LLM-evaluated metrics have been shown to be consistent with human judgments, as discussed in Sec.~\ref{sec:online}.

\vspace{-2mm}
\subsection{Overall Results}\label{sec:main_exp}

We consider the following baselines:
\begin{itemize}[leftmargin=*]
    \item IP \cite{gunawan2014mathematical}: A traditional IP method. We simplify it to use LLM for time budgeting and to consider POI ratings as utilities.
    \item Ernie-Bot 4.0 \cite{sun2021ernie}: The best-performing model on Chinese LLM tasks, selected as our dataset and system are in Chinese.
    \item GPT-3.5, GPT-4 and GPT-4 CoT \cite{openai2023gpt}: ChatGPT models with or without Chain-of-Thought~\cite{wei2022chain}.
    % \item TravelPlanner~\cite{Xie2024TravelPlanner}: Although it is a benchmark for LLMs in itinerary planning without direct solutions, we ran its two-stage method to generate itineraries using GPT-4. Only the "attractions" (POIs) generated were considered.
\end{itemize}
The baseline IP and our method do not compute the Fail Rate since the candidate POIs are all from the dataset. %  \todo{and TravelPlanner?}

\noindent The result is shown in Tab.~\ref{tab:main-exp}. 
% Notably, TravelPlanner does not account for spatial constraints on the attractions, even though latitude and longitude data are provided. While TravelPlanner is designed with tools to gather information, it is not optimized for generating fine-grained, spatially coherent itineraries. 
% Our proposed \textsc{ItiNera} outperforms all baselines across all metrics and achieves better or comparable results compared with ground truth data. 
% It shows a $\approx$12\% improvement in Recall Rate over TravelPlanner and generates itineraries only $\approx$100 meters longer per POI than the shortest TSP-solved path. 
% % \textsc{ItiNera} is also the only method to outperform GPT-4 CoT in LLM-evaluated metrics, especially in Match. 
% These results highlight \textsc{ItiNera}'s effectiveness in enhancing spatial coherence and aligning with user requests in OUIP.
our proposed \textsc{ItiNera} outperforms all baselines across all metrics and achieves better or comparable results compared with ground truth data. It shows a $\approx$30\% improvement in rule-based metrics over the best baseline, demonstrating superior personalization of user experiences. It maintains spatial coherence, generating itineraries only $\approx$100 meters longer per POI than the shortest TSP-solved path. \textsc{ItiNera} is also the only method to outperform GPT-4 CoT in LLM-evaluated metrics, especially in Match. These results highlight \textsc{ItiNera}'s effectiveness in enhancing spatial coherence and aligning with user requests in OUIP.

\begin{table*}[!h]
\centering
\resizebox{0.9\linewidth}{!}{
    \small
    \begin{tabular}{@{}lcc|ccc|ccc|ccc@{}}
    \toprule
    \multirow{2}{*}{Method} & \multicolumn{2}{c|}{Rule-based} & \multicolumn{3}{c|}{POI Quality} & \multicolumn{3}{c|}{Itinerary Quality} & \multicolumn{3}{c}{Match} \\ \cmidrule(l){2-12} 
     & AM & OL & Expert & User & LLM & Expert & User & LLM & Expert & User & LLM \\ \midrule
    GPT-4 CoT & 511.4 & 0.79 & 3.2 & 3.6 & 30\% & 2.5 & 3.0 & 32\% & 2.9 & 2.6 & 28\% \\
    \textsc{ItiNera} & 107.6 & 0.44 & 3.8 & 4.3 & 70\% & 3.2 & 3.8 & 68\% & 3.6 & 3.5 & 72\% \\ \bottomrule
    \end{tabular}
}
\vspace{-0.5em}
\caption{Deployed System Performance.}
\vspace{-1em}
\label{tab:subjective_evaluation}
\end{table*}

\subsection{Ablation Study}\label{sec:ablation}

To validate the effect of each component, we compare the following variants of \textsc{ItiNera}:
\begin{itemize}[leftmargin=*]
    \item GPT-4 CoT + UPC: integrates the UPC module to LLMs to generate itineraries based on user-owned POIs.
    \item \textsc{ItiNera} w/o RD: uses the entire user input string's embedding to retrieve POIs.
    \item \textsc{ItiNera} w/o PPR: quantifies the contribution of the PPR module compared to our full system.
    \item \textsc{ItiNera} w/o CSO: removes the CSO module and lets the LLM in the IG module determine the order of candidate POIs for the final itineraries.
    \item \textsc{ItiNera} w/ GPT-3.5 or LLaMA 3.1 8B: replaces GPT-4 with either GPT-3.5 or LLaMA 3.1 8B for generating the final itinerary.
\end{itemize}
We remove Fail Rate in the ablation study since all variants equipped with UPC never generate POIs not present in the database.

The results in Tab.~\ref{tab:ablation} show that UPC enhances the Recall Rate and Match of the GPT-4 CoT baseline. Variants ``w/o RD,'' ``w/o PPR,'' and ``w/ GPT-3.5'' have lower Recall Rate, POI Quality, Itinerary Quality, and Match than our full model, indicating a trade-off between spatial optimization and alignment with user requests. This parallels other conditional generation tasks, where aligning with human preferences can reduce inherent system ability \cite{saharia2022photorealistic, di2021video}. Removing the CSO module worsens the Average Margin and Overlaps but improves Recall Rate, POI Quality, and Match, showing the full model balances alignment with spatial ability. ``W/o PPR'' shows that the PPR module can address LLM context window limitations and save costs. Finally, ``w/ GPT-3.5'' outperforms the GPT-3.5 baseline, demonstrating our system's adaptability to different LLMs.

To validate that our method is a general framework compatible with both open-source LLMs and commercial ones, we conduct experiments with LLaMA 3.1-8B-Instruct~\cite{dubey2024llama}, a state-of-the-art model suitable for consumer GPUs.
LLaMA 3.1 offers performance comparable to GPT-3.5. Nevertheless, the performance of open-source models still lags behind commercial models. Considering the maintenance cost of hosting open-source models locally, we opt to use commercial models through API following most companies.

% While using commercial models certainly induces a cost, it is unclear if that would be more costly than the GPU cost to host open-source models (and the associated maintenance cost), especially models at a size that affords similar capability as state-of-the-art commercial models. Indeed, many current companies opt to use commercial models as a part of their business, sometimes with credits from the model providers, which we follow. In particular, the API cost has been drastically decreasing over recent months/years, and we expect this to continue and to make our proposed tool more scalable.\todo{revise this paragraph}

\vspace{-1mm}
\subsection{Deployed System Performance}\label{sec:online}
\label{sec:subjective_evaluation}
\vspace{-1mm}

Our deployed system is currently accessible to a select group of users recommended by our partnered travel agency.
To verify the effectiveness of our system in real-world scenarios, we conduct human evaluations.
Human evaluation has been extensively employed in prior research on generative tasks \cite{saharia2022photorealistic,rombach2022high,zhuo2023video} where objective metrics fail to adequately assess specific dimensions of output quality.
We invite 464 regular users of our system (\textit{User}) and 33 experienced travel assistants from our partnered travel agency (\textit{Expert}) to compare the two itineraries (randomly ordered) generated by GPT-4 CoT and our system based on their requests. 

The average evaluation results in Tab.~\ref{tab:subjective_evaluation} show that our method is preferred by both experts and regular users across all metrics, especially for Match, validating the effectiveness of our system in real-world scenarios. The human evaluation results are consistent with the LLM evaluation win rate, indicating that the proposed LLM-evaluated metrics are appropriate and adaptable 
when rule-based evaluation is insufficient.

\subsection{Qualitative Results} \label{supple:qualitative}

\begin{figure}[h]
    \centering
    \includegraphics[width=\linewidth]{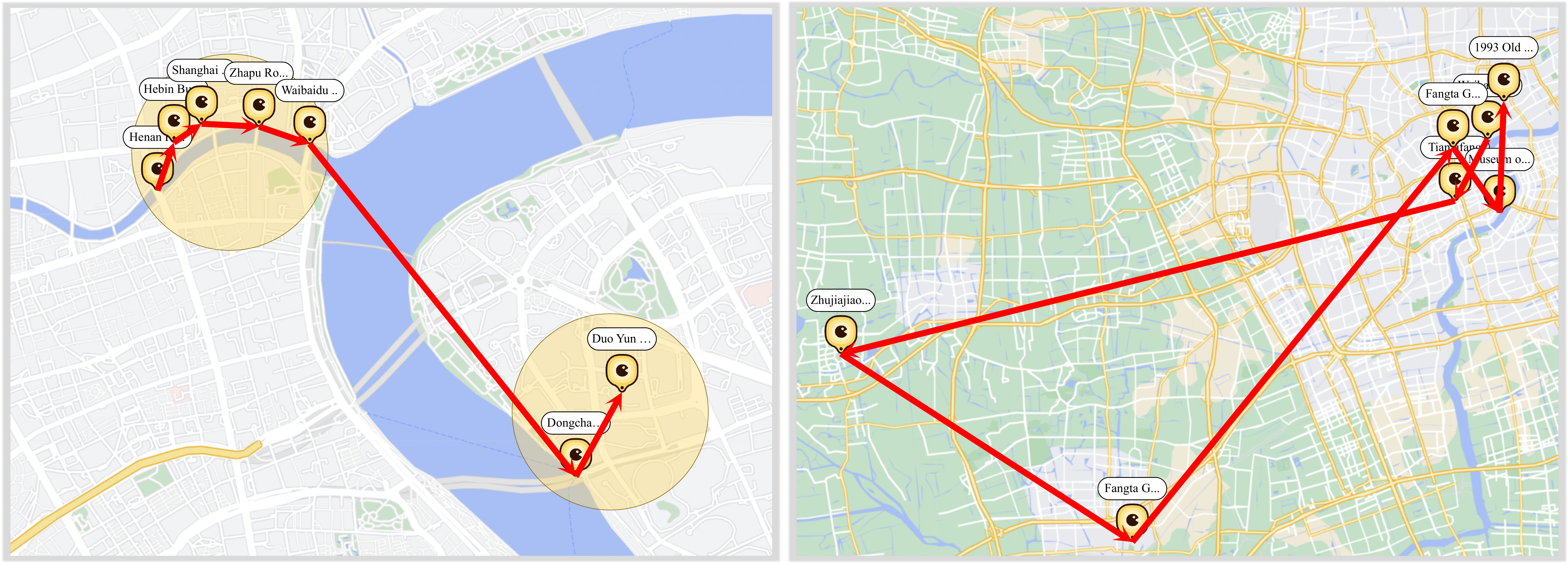}
    \vspace{-1.5em}
    \caption{Generated itineraries of \textsc{ItiNera} (left) and GPT-4 CoT (right).}
    \vspace{-0.5em}
    \label{fig:exp_qualitative}
\end{figure}

We further conduct a qualitative study to demonstrate the importance of integrating LLM with spatial optimization.
Consider a user request ``I'm seeking an artsy itinerary that includes exploring the river's bridges and ferries'', we visualize the results from \textsc{ItiNera} and GPT-4 CoT in Fig. \ref{fig:exp_qualitative}.

We find that our itinerary better matches the user preferences. The itinerary passes several bridges along the Huangpu River, includes a ferry crossing, and concludes at the art-atmosphere-rich Duoyun Bookstore, offering a restful endpoint for users. In contrast, the POIs selected by GPT are more mainstream.
Moreover, our spatial arrangement is more logical, avoiding detours and concentrating selected POIs within two spatial clusters. The itinerary generated by GPT is spatially poor, has a disordered sequence of visits, and contains excessively distant POIs. Beyond this example, GPT also risks hallucinating non-existent POIs, highlighting the superiority of our system in comparison.

\vspace{-1mm}
\section{Conclusion}
\vspace{-1mm}
We introduce the OUIP problem and a solution \textsc{ItiNera} that integrates LLMs with spatial optimization. \textsc{ItiNera} enables the generation of personalized and spatially coherent itineraries directly from natural language requests.  
Experiments on the real-world dataset and deployed system performance validate the effectiveness of our approach.
This study not only sets a new benchmark for itinerary planning technologies but also broadens venues for further innovations in leveraging LLMs for complex problem-solving in urban contexts.

% \vspace{-1mm}
% \section*{Acknowledgement}
% \vspace{-1mm}

% We thank Han Zheng at MIT and Tiange Luo at UMich for participating in discussions about the initial idea and providing valuable insights.

\section*{Limitations}
\vspace{-1mm}
Despite the success of \textsc{ItiNera} in generating personalized itineraries, our system has several limitations. First, while the spatial optimization module works well in many cases, it may face efficiency challenges in highly complex urban environments. Moreover, although LLMs provide significant language processing capabilities, they still exhibit limitations in spatial reasoning and real-time decision-making, which may impact the quality of the generated itineraries in specific scenarios.

\section*{Ethical Statement}
\vspace{-1mm}
This study adheres to strict ethical guidelines to protect user privacy and data. The personalized POI database is user-owned, and all data processing follows legal data security and user consent standards. We have designed the system to be fair and inclusive, avoiding biases in itinerary recommendations across diverse user groups. Additionally, we emphasize environmental responsibility by ensuring that the system promotes sustainable urban tourism without adversely affecting local culture or ecosystems. Finally, transparency in the pipeline is a priority, ensuring users understand how their itineraries are generated.

\bibliography{final}

\begin{thebibliography}{33}
\providecommand{\natexlab}[1]{#1}

\bibitem[{Bolzoni et~al.(2014)Bolzoni, Helmer, Wellenzohn, Gamper, and Andritsos}]{bolzoni2014efficient}
Paolo Bolzoni, Sven Helmer, Kevin Wellenzohn, Johann Gamper, and Periklis Andritsos. 2014.
\newblock Efficient itinerary planning with category constraints.
\newblock In \emph{Proceedings of the 22nd ACM SIGSPATIAL international conference on advances in geographic information systems}, pages 203--212.

\bibitem[{Chen et~al.(2013)Chen, Wu, Zhou, and Tung}]{chen2013automatic}
Gang Chen, Sai Wu, Jingbo Zhou, and Anthony~KH Tung. 2013.
\newblock Automatic itinerary planning for traveling services.
\newblock \emph{IEEE transactions on knowledge and data engineering}, 26(3):514--527.

\bibitem[{de~Zarz{\`a} et~al.(2023)de~Zarz{\`a}, de~Curt{\`o}, Roig, and Calafate}]{de2023llm}
I~de~Zarz{\`a}, J~de~Curt{\`o}, Gemma Roig, and Carlos~T Calafate. 2023.
\newblock Llm multimodal traffic accident forecasting.
\newblock \emph{Sensors}, 23(22):9225.

\bibitem[{Di et~al.(2021)Di, Jiang, Liu, Wang, Zhu, He, Liu, and Yan}]{di2021video}
Shangzhe Di, Zeren Jiang, Si~Liu, Zhaokai Wang, Leyan Zhu, Zexin He, Hongming Liu, and Shuicheng Yan. 2021.
\newblock Video background music generation with controllable music transformer.
\newblock In \emph{Proceedings of the 29th ACM International Conference on Multimedia}, pages 2037--2045.

\bibitem[{Dubey et~al.(2024)Dubey, Jauhri, Pandey, Kadian, Al-Dahle, Letman, Mathur, Schelten, Yang, Fan et~al.}]{dubey2024llama}
Abhimanyu Dubey, Abhinav Jauhri, Abhinav Pandey, Abhishek Kadian, Ahmad Al-Dahle, Aiesha Letman, Akhil Mathur, Alan Schelten, Amy Yang, Angela Fan, et~al. 2024.
\newblock The llama 3 herd of models.
\newblock \emph{arXiv preprint arXiv:2407.21783}.

\bibitem[{Germano(2023)}]{Germano_2023}
Antonello Germano. 2023.
\newblock \href {https://daxueconsulting.com/citywalk/} {Citywalk: embracing urban charms and captivating generation z}.

\bibitem[{Gionis et~al.(2014)Gionis, Lappas, Pelechrinis, and Terzi}]{gionis2014customized}
Aristides Gionis, Theodoros Lappas, Konstantinos Pelechrinis, and Evimaria Terzi. 2014.
\newblock Customized tour recommendations in urban areas.
\newblock In \emph{Proceedings of the 7th ACM international conference on Web search and data mining}, pages 313--322.

\bibitem[{Gunawan et~al.(2014)Gunawan, Yuan, and Lau}]{gunawan2014mathematical}
Aldy Gunawan, Zhi Yuan, and Hoong~Chuin Lau. 2014.
\newblock A mathematical model and metaheuristics for time dependent orienteering problem.
\newblock PATAT.

\bibitem[{Halder et~al.(2024)Halder, Lim, Chan, and Zhang}]{halder2024survey}
Sajal Halder, Kwan~Hui Lim, Jeffrey Chan, and Xiuzhen Zhang. 2024.
\newblock A survey on personalized itinerary recommendation: From optimisation to deep learning.
\newblock \emph{Applied Soft Computing}, 152:111200.

\bibitem[{Ho and Lim(2022)}]{ho2022poibert}
Ngai~Lam Ho and Kwan~Hui Lim. 2022.
\newblock Poibert: A transformer-based model for the tour recommendation problem.
\newblock In \emph{2022 IEEE International Conference on Big Data (Big Data)}, pages 5925--5933. IEEE.

\bibitem[{Hsueh and Huang(2019)}]{hsueh2019personalized}
Yu-Ling Hsueh and Hong-Min Huang. 2019.
\newblock Personalized itinerary recommendation with time constraints using gps datasets.
\newblock \emph{Knowledge and Information Systems}, 60(1):523--544.

\bibitem[{Jiang et~al.(2014)Jiang, Gao, Li, Wu, and Pei}]{jiang2014hierarchical}
Jingqing Jiang, Jingying Gao, Gaoyang Li, Chunguo Wu, and Zhili Pei. 2014.
\newblock Hierarchical solving method for large scale tsp problems.
\newblock In \emph{International symposium on neural networks}, pages 252--261. Springer.

\bibitem[{Mo et~al.(2023)Mo, Xu, Zhuang, Ma, Guo, and Zhao}]{mo2023large}
Baichuan Mo, Hanyong Xu, Dingyi Zhuang, Ruoyun Ma, Xiaotong Guo, and Jinhua Zhao. 2023.
\newblock Large language models for travel behavior prediction.
\newblock \emph{arXiv preprint arXiv:2312.00819}.

\bibitem[{OpenAI(2023)}]{openai2023gpt}
OpenAI. 2023.
\newblock Gpt-4 technical report.
\newblock \emph{arXiv preprint arXiv:2303.08774}.

\bibitem[{Rani et~al.(2018)Rani, Kholidah, and Huda}]{rani2018development}
Septia Rani, Kartika~Nur Kholidah, and Sheila~Nurul Huda. 2018.
\newblock A development of travel itinerary planning application using traveling salesman problem and k-means clustering approach.
\newblock In \emph{Proceedings of the 2018 7th International Conference on Software and Computer Applications}, pages 327--331.

\bibitem[{Rombach et~al.(2022)Rombach, Blattmann, Lorenz, Esser, and Ommer}]{rombach2022high}
Robin Rombach, Andreas Blattmann, Dominik Lorenz, Patrick Esser, and Bj{\"o}rn Ommer. 2022.
\newblock High-resolution image synthesis with latent diffusion models.
\newblock In \emph{Proceedings of the IEEE/CVF conference on computer vision and pattern recognition}, pages 10684--10695.

\bibitem[{Saharia et~al.(2022)Saharia, Chan, Saxena, Li, Whang, Denton, Ghasemipour, Gontijo~Lopes, Karagol~Ayan, Salimans et~al.}]{saharia2022photorealistic}
Chitwan Saharia, William Chan, Saurabh Saxena, Lala Li, Jay Whang, Emily~L Denton, Kamyar Ghasemipour, Raphael Gontijo~Lopes, Burcu Karagol~Ayan, Tim Salimans, et~al. 2022.
\newblock Photorealistic text-to-image diffusion models with deep language understanding.
\newblock \emph{Advances in Neural Information Processing Systems}, 35:36479--36494.

\bibitem[{Sun et~al.(2021)Sun, Wang, Feng, Ding, Pang, Shang, Liu, Chen, Zhao, Lu et~al.}]{sun2021ernie}
Yu~Sun, Shuohuan Wang, Shikun Feng, Siyu Ding, Chao Pang, Junyuan Shang, Jiaxiang Liu, Xuyi Chen, Yanbin Zhao, Yuxiang Lu, et~al. 2021.
\newblock Ernie 3.0: Large-scale knowledge enhanced pre-training for language understanding and generation.
\newblock \emph{arXiv preprint arXiv:2107.02137}.

\bibitem[{Sylejmani et~al.(2017)Sylejmani, Dorn, and Musliu}]{sylejmani2017planning}
Kadri Sylejmani, J{\"u}rgen Dorn, and Nysret Musliu. 2017.
\newblock Planning the trip itinerary for tourist groups.
\newblock \emph{Information Technology \& Tourism}, 17:275--314.

\bibitem[{Tang et~al.(2022)Tang, He, and Zhao}]{tang2022hgarn}
Yihong Tang, Junlin He, and Zhan Zhao. 2022.
\newblock Hgarn: Hierarchical graph attention recurrent network for human mobility prediction.
\newblock \emph{arXiv preprint arXiv:2210.07765}.

\bibitem[{Tang et~al.(2024)Tang, Qu, Wang, Zhuang, Wu, Ma, Wang, Zheng, Zhao, and Zhao}]{tang2024sparkle}
Yihong Tang, Ao~Qu, Zhaokai Wang, Dingyi Zhuang, Zhaofeng Wu, Wei Ma, Shenhao Wang, Yunhan Zheng, Zhan Zhao, and Jinhua Zhao. 2024.
\newblock Sparkle: Mastering basic spatial capabilities in vision language models elicits generalization to composite spatial reasoning.
\newblock \emph{arXiv preprint arXiv:2410.16162}.

\bibitem[{Taylor et~al.(2018)Taylor, Lim, and Chan}]{taylor2018travel}
Kendall Taylor, Kwan~Hui Lim, and Jeffrey Chan. 2018.
\newblock Travel itinerary recommendations with must-see points-of-interest.
\newblock In \emph{Companion Proceedings of the The Web Conference 2018}, pages 1198--1205.

\bibitem[{Wang et~al.(2023)Wang, Fang, Zeng, and Cheng}]{wang2023would}
Xinglei Wang, Meng Fang, Zichao Zeng, and Tao Cheng. 2023.
\newblock Where would i go next? large language models as human mobility predictors.
\newblock \emph{arXiv preprint arXiv:2308.15197}.

\bibitem[{Wei et~al.(2022)Wei, Wang, Schuurmans, Bosma, Xia, Chi, Le, Zhou et~al.}]{wei2022chain}
Jason Wei, Xuezhi Wang, Dale Schuurmans, Maarten Bosma, Fei Xia, Ed~Chi, Quoc~V Le, Denny Zhou, et~al. 2022.
\newblock Chain-of-thought prompting elicits reasoning in large language models.
\newblock \emph{Advances in Neural Information Processing Systems}, 35:24824--24837.

\bibitem[{Xie et~al.(2024)Xie, Zhang, Chen, Zhu, Lou, Tian, Xiao, and Su}]{Xie2024TravelPlanner}
Jian Xie, Kai Zhang, Jiangjie Chen, Tinghui Zhu, Renze Lou, Yuandong Tian, Yanghua Xiao, and Yu~Su. 2024.
\newblock Travelplanner: A benchmark for real-world planning with language agents.
\newblock \emph{arXiv preprint arXiv: 2402.01622}.

\bibitem[{Xue et~al.(2022)Xue, Voutharoja, and Salim}]{xue2022leveraging}
Hao Xue, Bhanu~Prakash Voutharoja, and Flora~D Salim. 2022.
\newblock Leveraging language foundation models for human mobility forecasting.
\newblock In \emph{Proceedings of the 30th International Conference on Advances in Geographic Information Systems}, pages 1--9.

\bibitem[{Yan et~al.(2023)Yan, Wen, Zhong, Chen, Chen, Wen, Zimmermann, and Liang}]{yan2023urban}
Yibo Yan, Haomin Wen, Siru Zhong, Wei Chen, Haodong Chen, Qingsong Wen, Roger Zimmermann, and Yuxuan Liang. 2023.
\newblock When urban region profiling meets large language models.
\newblock \emph{arXiv preprint arXiv:2310.18340}.

\bibitem[{Yochum et~al.(2020)Yochum, Chang, Gu, and Zhu}]{yochum2020adaptive}
Phatpicha Yochum, Liang Chang, Tianlong Gu, and Manli Zhu. 2020.
\newblock An adaptive genetic algorithm for personalized itinerary planning.
\newblock \emph{IEEE Access}, 8:88147--88157.

\bibitem[{Zhang and Tang(2018)}]{zhang2018itinerary}
Yu~Zhang and Jiafu Tang. 2018.
\newblock Itinerary planning with time budget for risk-averse travelers.
\newblock \emph{European Journal of Operational Research}, 267(1):288--303.

\bibitem[{Zheng et~al.(2023{\natexlab{a}})Zheng, Abdel-Aty, Wang, Wang, and Ding}]{zheng2023trafficsafetygpt}
Ou~Zheng, Mohamed Abdel-Aty, Dongdong Wang, Chenzhu Wang, and Shengxuan Ding. 2023{\natexlab{a}}.
\newblock Trafficsafetygpt: Tuning a pre-trained large language model to a domain-specific expert in transportation safety.
\newblock \emph{arXiv preprint arXiv:2307.15311}.

\bibitem[{Zheng et~al.(2023{\natexlab{b}})Zheng, Wang, Wang, and Ding}]{zheng2023chat}
Ou~Zheng, Dongdong Wang, Zijin Wang, and Shengxuan Ding. 2023{\natexlab{b}}.
\newblock Chat-gpt is on the horizon: Could a large language model be suitable for intelligent traffic safety research and applications?
\newblock \emph{ArXiv}.

\bibitem[{Zhou et~al.(2024)Zhou, Lin, and Li}]{zhou2024large}
Zhilun Zhou, Yuming Lin, and Yong Li. 2024.
\newblock Large language model empowered participatory urban planning.
\newblock \emph{arXiv preprint arXiv:2402.01698}.

\bibitem[{Zhuo et~al.(2023)Zhuo, Wang, Wang, Liao, Bao, Peng, Han, Zhang, Fang, and Liu}]{zhuo2023video}
Le~Zhuo, Zhaokai Wang, Baisen Wang, Yue Liao, Chenxi Bao, Stanley Peng, Songhao Han, Aixi Zhang, Fei Fang, and Si~Liu. 2023.
\newblock Video background music generation: Dataset, method and evaluation.
\newblock In \emph{Proceedings of the IEEE/CVF International Conference on Computer Vision}, pages 15637--15647.

\end{thebibliography}

\clearpage

\appendix

\onecolumn

\section{Demonstration of the Deployed System}\label{supple:demo_deploy}

\begin{figure}[h]
    \centering
    \includegraphics[width=.75\linewidth]{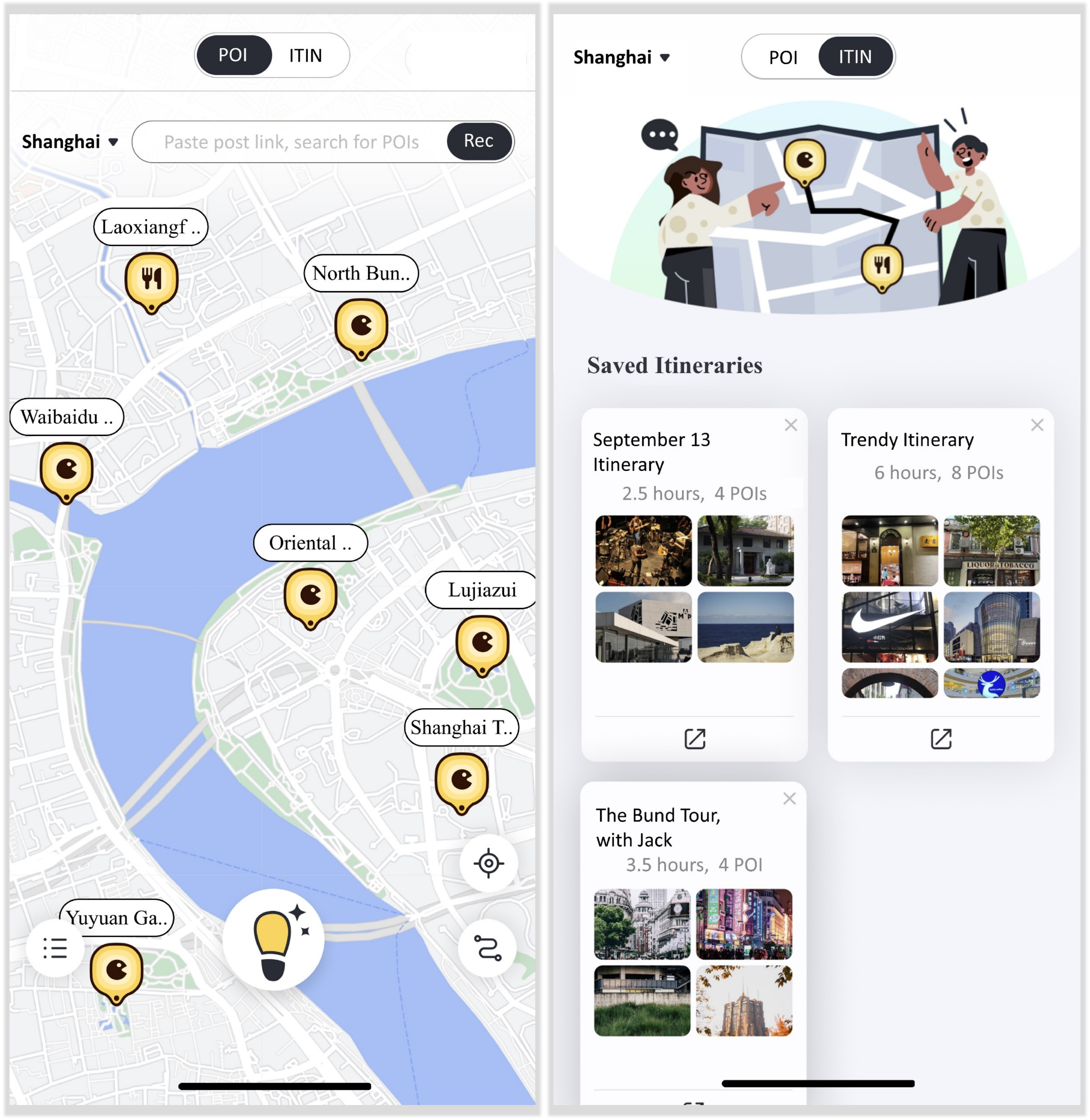}
    \caption{Screenshots of the deployed system: POI view \& Itinerary view.}
    \label{fig:deployed}
\end{figure}

We provide screenshots of our deployed system in Fig.~\ref{fig:deployed}. The left screenshot shows the POI interface, where users can add new POIs by direct searching or pasting a link of a travel-related post. They can filter their desired POIs to display on their personal map. The POI icon represents its category (entertainment, restaurant, etc.). Users can select several POIs by pressing the bottom right button to create an itinerary. They can also use our system to generate an itinerary from natural language input (the left figure of Fig.~\ref{fig:itiplan}).

The right screenshot shows the itinerary interface. Users can browse the itineraries they have created and generated. They can tap one itinerary to see the details (the middle and right figures of Fig.~\ref{fig:itiplan}).

\clearpage

\section{Dataset} \label{supple:dataset}

In this section, we provide the data format of the collected real-world dataset. Specifically, the data for each city contains two tables: one is the POI table, which primarily stores the POIs and their features, and the other is the Itinerary table, which is used to store users' natural language requests and the corresponding ground truth itineraries. 
\begin{table}[h]
\centering 
\resizebox{\textwidth}{!}{
\begin{tabular}{llllllllll}
\hline
id & name & address & city & description & longitude & latitude & rating & category & context \\
1 & The Bund & Zhongshan East 1st Rd, Huangpu & Shanghai & The Bund is a waterfront area ... & 121.4906033011 & 31.2377704249 & 5.0 & site & ... \\ \hline
\end{tabular}
}
\caption{POI data sample.}
 \label{data:poisample}
\end{table}

The sample POI data is shown in Table \ref{data:poisample}, where the context column is a concatenation of the strings from all the previous columns. The embedding of each POI is also obtained by calling $E_{\theta^\prime}$ to embed the context field. The resulting embedding, $\mathcal{E}$, contains rich semantic information about the POIs.

\begin{table}[h]
\centering
\begin{tabular}{ll}
\hline
user\_request & itinerary \\
I'm seeking an artsy itinerary that includes exploring the river's bridges and ferries. & {[}1, 3, 6, ...{]} \\ \hline
\end{tabular}
\caption{Itinerary data sample.}
 \label{data:itisample}
\end{table}

The sample itinerary data is shown in Table \ref{data:itisample}, which contains two columns: one for the user's request and the other storing a list of POI IDs representing the ground truth itinerary (label) for the user's request.

\vspace{3em}
\section{Implementation Details} \label{supple:imple}

\subsection{Method Implementation}

We use the OpenAI text-embedding-ada-002 model for embedding purposes. The spatial coherence of itineraries is optimized through an open-source TSP solver\footnote{\url{https://github.com/fillipe-gsm/python-tsp}}. Integration of POI data, including geographical coordinates, user ratings, categorizations, and physical addresses, is facilitated through the Amap API\footnote{\url{https://lbs.amap.com/}}.

\subsection{Baseline Settings}
We use the same itinerary generation prompt for all baselines, including basic task requirements and output format, as in $\mathbb{P}_{IG}$ in \S\ref{supple:prompt_ig}. For GPT-4 CoT, we extend the prompt by integrating ``thoughts'', detailed in \S\ref{supple:prompt_baseline}.

We adopt the $\mathbb{P}_{IT}$ for the baseline IP for time budgeting. We prompt the LLM baselines to generate itineraries based on user requests. We searched for each POI in the generated itinerary using the Map API. Here, the database associated with the Map API is considered to be the current collection of all existing POIs. We leverage fuzzy string matching\footnote{\url{https://github.com/seatgeek/thefuzz}} to determine if there is a match with specific POIs. The failed POIs contribute to the failure rate metrics. For the matched POIs, attributes of the POI (such as location) are attached to the itinerary for subsequent evaluation.

\clearpage

\section{Cluster-aware Spatial Optimization Supplementary}\label{supple:spatial}
We present the details of the implementation of algorithms involved in cluster-aware spatial optimization. 

\subsection{SolveTSP}

\begin{algorithm}
{
\captionsetup{font=normalsize}
\caption{Simulated Annealing for TSP}\label{alg:sa_tsp}
}
\begin{algorithmic}[1]
\Procedure{SimulatedAnnealing}{$\textbf{cities}$, $T_{init}$, $T_{min}$, $\alpha$}
    \State $solution \gets \text{RandomSolution}(cities)$
    \State $T \gets T_{init}$
    \While{$T > T_{min}$}
        \State $newSolution \gets \text{Neighbor}(solution)$
        \State $costDifference \gets \text{Cost}(newSolution) - \text{Cost}(solution)$
        \If{$costDifference < 0 \text{ or } \exp(-costDifference / T) > \text{Random}()$}
            \State $solution \gets newSolution$
        \EndIf
        \State $T \gets \alpha \times T$
    \EndWhile
    \State \textbf{return} $solution$
\EndProcedure
\end{algorithmic}
\end{algorithm}

`SolveTSP' implements a simulated annealing algorithm for efficiently solving the TSP problem with a large set of candidates. Simulated annealing is a classic metaheuristic approach where the model iteratively proposes a new solution and replaces the current solution if a certain condition is satisfied until the temperature goes down to zero. We detail the implementation for simulated annealing in Algo \ref{alg:sa_tsp}.

\begin{itemize}[leftmargin=*]
    \item \textsc{RandomSolution}: Generates a random permutation of the cities as the initial solution.
    \item \textsc{Neighbor}: Produces a new solution by making a small change to the current solution. In our implementation, we consider four types of operations including swapping two randomly selected cities, inverting a subroute, inserting a randomly selected city to another position, and inserting a randomly selected subroute to another position.
    \item \textsc{Cost}: Calculates the total distance of the proposed solution's path.
    \item \textbf{$T_{init}$} and \textbf{$T_{min}$}: The initial and minimum temperatures for the SA algorithm. In our implementation, \textbf{$T_{init}$} is set to 5000 and \textbf{${T_{min}}$} is set to 0. 
    \item \textbf{$\alpha$}: The cooling rate that determines how fast the temperature decreases. In our implementation, \textbf{$\alpha$} is set to 0.99.
\end{itemize}

\clearpage

\subsection{SolveTSPWithEndpoints}

In each cluster, the dataset typically comprises a limited set of candidate points. Consequently, the prioritization shifts towards optimizing the accuracy of the resultant solution rather than focusing solely on computational efficiency. To address the Traveling Salesman Problem (TSP) with predetermined starting and ending points, we adopt a linear programming (LP) methodology. We detail the formulation of the linear program in Alg. \ref{alg:ep_tsp}.

\begin{algorithm}
{
\captionsetup{font=normalsize}
\caption{SolveTSPWithEndpoints}\label{alg:ep_tsp}
}
\begin{algorithmic}[1]
\Require $dist$, $start\_point$, $end\_point$
\State \textbf{Solve the following linear program:}
\begin{align*}
\text{Minimize:} \quad & \min \sum_{i \neq j} x_{ij} \cdot dist[i][j] \\
& \textcolor{darkgray}{// \text{Ensures each internal node in optimal path has in-degree 1 and out-degree 1}}  \\
\text{Subject to:} \quad 
& \sum_{i \neq k} x_{ik} = 1, \quad \forall k \neq s, e \\
& \sum_{i \neq k} x_{ki} = 1, \quad \forall k \neq s, e \\
& \textcolor{darkgray}{// \text{Add constraints for source node and sink node}}  \\
& \sum_{i \neq s} x_{si} = 1 \\
& \sum_{i \neq s} x_{is} = 0 \\
& \sum_{i \neq e} x_{ie} = 1 \\
& \sum_{i \neq e} x_{ei} = 0 \\
& \textcolor{darkgray}{// \text{Eliminates all subtours}}  \\
& \sum_{i \in S} \sum_{j \notin S, j \neq i} x_{ij} \leq |S| - 1, \quad \forall S \subset \{1, \ldots, n\}, S \neq \emptyset, S \neq \{1, \ldots, n\} \\
& \textcolor{darkgray}{// \text{Add binary variable constraints}}  \\
& x_{ij} \in \{0, 1\}, \quad \forall i, j
\end{align*}
\State \textbf{Return} Optimal path
\end{algorithmic}
\end{algorithm}

\clearpage

% \section{Experiments on More Cities}\label{supple:moreexp}

% We provide extended experiments on two more subsets of large and small cities of our dataset: Beijing and Hangzhou. As shown in Tab.~\ref{tab:supp_main}, the results are similar to Sec.~\ref{sec:main_exp}.

% \input{Tables_arxiv//supp_main_exp}

\vspace{1cm}
\section{Overview of the POI extraction pipeline}\label{supple:pipe}

\vspace{10pt}
\begin{tcolorbox}[title=Prompt for Extracting POI Names and Locations, breakable]
\begin{minted}{markdown}
# Guidelines

## Task Background
Your task is to identify and extract mentioned cations in the posts/travelogues provided by users to help them quickly find these places on a map. Now, based on the content of the post and in context, carry out the extraction and description of Points of Interest (POIs) mentioned in the post. Focus primarily on places that can be visited, rather than merely on place names.

## Notes on Handling POIs
1. Comprehensive Definition of POI: Typically used to describe a specific geographical location or site, such as restaurants, hotels, streets, attractions, museums, bars, cafes, malls, etc. These locations or sites may have specific value or interest to users or travelers.
2. Characteristics of POI: Specific places recommended or mentioned in the post that are usable for dining, entertainment, etc.
3. Specificity: A POI refers to a specific, particular place, not a broad geographical area or city name.
4. Uniqueness: When a text is separated by symbols like "/", "&", ",", for example, "Julu Road/Tianzifang", it often represents two POIs, in this case, "Julu Road" and "Tianzifang" should be extracted separately.
5. Examples of POI: Specific restaurants, performance venues, attractions, shops, streets, etc.
6. Non-POI Examples: Collections of places, food names, types of cuisine, performance groups, exhibition events, etc.

## Post Structure
Title: The post's title.
Text: The main body content of the post.
Text in the images: text recognized from the images.
Transcribed text: text transcribed from the video.

## Task Process
1. Extraction: Based on your reasoning, judgment, and knowledge, extract all mentioned POIs from the post.
2. Verification: In the context of each POI, ensure all POIs fit the definition and are specific places.
3. Address Information: In the context of each POI, find related address information that can be searched on a map, such as "158 Julu Road, Shanghai."
4. Handling No Information: If no location information is available, return an empty POI list: {{}}.
5. Formatting: Organize information into the specified JSON structure.

## Output Format

### Specific Format

{{
    "POI Name": "Related Address Information for the POI"
}}

### Examples

Example 1:
If the original post mentions "Lao Nong Tang Noodle Shop in Luxi: A time-honored noodle shop that appears on Shanghai's must-eat list all year round!", the output for this POI should be
{{
"Lao Nong Tang Noodle Shop in Luxi": null
}}

Example 2:
If the original post mentions "Red Baron (Jianye District Wentiyi Road branch)
Looking around, the most striking red on the entire Wentiyi Road, seamlessly blending with Mixue Bingcheng", the output for this POI should be
{{
"Red Baron (Jianye District Wentiyi Road branch)": null
}}

## Output Standards

- The output is a dictionary, with keys being the POI names and values being the related address information for the POI. If address information is missing, please use "null" to fill in.
- Ensure the output is in valid JSON format and can be parsed by Python json.loads.


## Task Start

Please begin processing the post content: ```{post_info}```.

Note: Ensure the return format follows {{Point of Interest Name: Related Address Information}}. Ensure it can be json.loads parsed.

\end{minted}
\end{tcolorbox}

The prompt for extracting POI names and locations is provided above. As illustrated in Sec.~\ref{sec:ppa}, we design a pipeline to automatically extract POIs and relevant information from user-generated content on various social media platforms. The pipeline consists of the following steps:

\begin{itemize}[leftmargin=*]
    \item Scrape text, images, and videos from the input link of a travel-related post.
    \item Use automatic speech recognition to obtain transcription from the video and optical character recognition (OCR) to extract text from the images. Merge them with the original text to obtain the post information.
    \item Use GPT-3.5 to extract POI names and locations from the post information.
    \item Use map service API to look up the extracted POI names, obtain the coordinates, and extract POI names, similar to the evaluation pipeline in Sec.\ref{supple:imple}.
    \item Use GPT-3.5 to generate POI descriptions from the POI names and post information.
\end{itemize}

The prompt for generating POI descriptions is provided below.

\begin{tcolorbox}[title=Prompt for Generating POI Descriptions, breakable]
\begin{minted}{markdown}
{post_info}

Based on the content of the above post, please write out the reasons for recommending each location to tourists in the following list. 
Consider what can be done at this location, its features, and why it is fun. 
If the original post lacks information, you may appropriately supplement based on your knowledge, but please ensure brevity. 
The related information for each point should not exceed 30 words. 
The results should be output in JSON format, specifically in the form {{Place Name: Information related to the place from the original post}}, where "Information related to the place from the original post" should be a sentence or phrase, like a string. 
If a place does not have any relevant information, fill in the description related to the place from the original post with "null".
\end{minted}
\end{tcolorbox}

We execute an automated process to extract POIs from the most recent trending posts and update a comprehensive POl database. At a regular interval of 24 hours, we obtain recent trending travel-related posts across multiple cities on social media platforms and run the above pipeline to extract POI names, locations, and descriptions to maintain the database.

\clearpage

\section{Prompts}

\subsection{Prompt for Decomposing User Requests}\label{supple:prompt_decom}

\begin{tcolorbox}[title=Prompt for Decomposing User Requests, breakable]
\begin{minted}{markdown}
Please help me break down a user request into multiple independent requirements, each including both positive and negative requirements. Return the results in the following format directly based on the **User Request** given, without writing any code.

### Output Format:

Return a list, where each item is a dictionary representing an independent requirement, with the following key-value pairs:
- **pos**: The positive requirement, representing what the user wants, excluding any negative requirements.
- **neg**: The negative requirement, generally what the user does not want, dislikes, or refuses. All negative requirements must be captured in this field, for example, "non-spicy" should extract "spicy", "don't want crowded places" should extract "crowded", "hate noisy" should extract "noisy".
- **mustsee**: Indicates whether this requirement represents a specific place name. If so, this field is `true`, otherwise, it is `false`.
- **type**: Indicates whether the requirement is for a "place" or an "itinerary", with place having sub-types "place", "starting point", and "ending point". Overall, this field can have the values "place", "starting point", "ending point", or "itinerary".

- Your return should be a list in the following format:
[
    {{
        "pos": "positive requirement", (excluding negative requirements)
        "neg": "negative requirement" (what's not wanted, disliked, refused, not wanting to go or see, any negated requirement),
        "mustsee": true (whether it's a must-see point, all specific places should be set to true),
        "type": "place" 
    }}, 
    ...
]
- The **positive requirement** must not be empty, and it must not include any negative requirements. All negative requirements should be summarized in the value of the "neg" field.
- Set to null in cases where there are no **negative requirements** for a specific place.
- Sometimes users only describe what they do not want (negative requirements), in such cases, you should summarize a **positive requirement** based on the **negative requirement**. For example, if a user says 'don't want spicy food', the output should include: "pos" corresponding to "food", "neg" corresponding to "spicy".
- Independent requirements must have specific descriptions or demands to be considered a requirement, for example, "recommend a route" does not count as an independent requirement.
- "mustsee" must be a specific place name, not a general term.
- If a place is definitively a "starting point" or "ending point", then the value of the "type" field should be "starting point" or "ending point", respectively. "Starting points" and "ending points" are must-see points, with the "mustsee" field set to true.
- A place can only be considered as a "starting point" or "ending point" if it is a specific attraction or location, and there can only be at most one "starting point" and one "ending point".
- The return should not include any other content.

### Example Outputs:

Example 1:
User Request: "I want to start by visiting Sinan Mansions, then find something fun to do nearby, and I don't want it to be crowded"
Output:
[
    {{
        "pos": "Sinan Mansions",
        "neg": null,
        "mustsee": true,
        "type": "starting point"
    }}, 
    {{
        "pos": "fun places near Sinan Mansions",
        "neg": "crowded",
        "mustsee": false,
        "type": "place"
    }}
]


### mustsee Field Assignment Examples
"mustsee" true for specific place names: "Hualian Supermarket", "Old Mac Coffee Shop", "Wukang Mansion", "Nanluoguxiang", ...
"mustsee" false for general place names: "supermarket", "milk tea shop", "bar", "coffee", ...

### Output Guidelines
- Return a list, each item in the list is a dictionary containing "pos", "neg", "mustsee", and "type" key-value pairs.
- Return as a JSON List.
- The list can be empty; if empty, just return a JSON list.
- The output should not include any other information, ensuring it can be parsed by json.loads.

### User Request
{user_req}


### Task Overview
Your task is to analyze and break down the **User Request** into independent requirements and return them.
1. First, separate the different independent requirements, breaking down each into positive and negative requirements.
2. Positive requirements should only include what the user wants, and negative requirements should only include what the user does not want.
3. For each independent requirement, refer to **mustsee field assignment examples** to assign a value to the "mustsee" field, analyzing whether the **positive requirement** is a specific place name. If so, set "mustsee" to true, otherwise set it to false.
4. Refer to the **examples** and **output format** to complete the other fields.

#### Notes:
- Do not include duplicate independent requirements; ensure each independent requirement corresponds to different key points in the user's needs.
- "Itinerary" requirements should be for the whole itinerary, such as including several places, approximate time, etc., all others are place requirements.
- The "type" field can only be one of ["place", "itinerary", "starting point", "ending point"].
- All attractions must be completely separated, such as "Nanluoguxiang and Drum Tower" must be split into "Nanluoguxiang" and "Drum Tower" as two requirements.

Now, based on the **user Request**, refer to the **example outputs**, and return according to the **output guidelines** and **output format**.
\end{minted}
\end{tcolorbox}

\subsection{Prompt for Indicating Travel Time of an Itinerary} \label{supple:prompt_time}

\begin{tcolorbox}[title=Prompt for Travel Time Indication, breakable]
\begin{minted}{markdown}
Please play the role of a top AI Travel Time Planning Assistant. Your job is to determine the time needed for a day's itinerary based on the user request. If the user request is empty, please default to ["4"].

## Task Overview
Your task is to return the required time for a day's itinerary based on given the user request. If the user request mention specific time constraints for the route, return the itinerary time directly as per the user's request, up to a maximum of 8 hours (return ["8"] if it exceeds 8 hours). Please return the itinerary time directly based on the user request, no need to write any code.

## User Request
{user_reqs}

## Input-Output Examples
- **Example 1**:
- **User Request**: "I'd like to visit a museum, enjoy authentic cuisine, and experience nightlife."
- **Output**: ["8"]

- **Example 2**:
- **User Request**: "I want to tour historical buildings and take in the city views“
- **Output**: ["6"]

- **Example 3** (In this example, the user specifies approximately **five hours** for the route):
- **User Request**: "I plan to explore the Huangpu River and Yu Garden for about five hours."
- **Output**: ["5"]

## Output Specifications
- Return a list of length 1, containing a single integer representing the required itinerary duration (in hours). The value range is 1 to 8.
- Return as a JSON List with only one element inside.
- The list can be empty; please only return a JSON list.
- Ensure your output contains no additional information and can be parsed by json.loads.

Now, based on the **User Request**, return the time required for a day's itinerary according to the **Output Specifications**.
\end{minted}
\end{tcolorbox}

In this work, we utilize the inference capability of LLMs to estimate the duration of an itinerary based on a user request, which is used to instruct the IG module to generate an itinerary with a reasonable duration. For more complicated considerations, such as stay duration and travel time between POIs, we leave them for future research.

\subsection{Prompt for Identifying the Start POI}

\begin{tcolorbox}[title=Prompt for Start POI Identification, breakable]
\begin{minted}{markdown}
Please act as a top-tier AI travel planning assistant. Your job is to return the index of the best starting point for a day trip itinerary based on user needs and provided candidate POIs. If the candidate POIs are empty, please default to returning ["0"].

### Task Overview
Your task is to return the index of the best starting point for an itinerary based on the given candidate POIs and the user request. Directly return the starting point's index based on user needs and candidate POIs, without writing any code.

### Candidate POIs
{candidate_strings}

### User Request
{user_reqs}

### Guidelines
1. Ensure the selected POI meets the user request.
2. The starting point should be close to its neighboring points.
3. Prioritize POIs like museums or art galleries, which usually require more exploration time.
4. Avoid starting from bars or clubs.

### Example Inputs and Outputs
- **Example 1**:
- **Candidate POIs**: ["Museum", "Park", "Bar"]
- **Output**: ["0"]

- **Example 2**:
- **Candidate POIs**: ["Shopping Center", "Art Gallery", "Historical Building"]
- **Output**: ["1"]

### Output Specification
1. Return a list of length 1, containing an integer that represents the index of the best starting point.
2. Return as a JSON List, with only one element inside.
3. The output should not contain any other information, ensuring it can be parsed by json.loads.
4. Your response should be a length-1 JSON list, where the index comes from {return_candidates}.
    - Example: ["0"]

Now, based on the **Candidate POIs** and **User Needs**, return the index of the best starting point for the day trip itinerary according to the **Output Specification**. Note, ensure your reply is **a list composed of a single number** from {return_candidates}, following the requirements in the **Output Specification** to return **a length-1 JSON list**, and do not return any other content.

\end{minted}
\end{tcolorbox}

% \vspace{20pt}

\clearpage

\subsection{Prompt for Generating the Itinerary} \label{supple:prompt_ig}

\begin{tcolorbox}[title=Prompt for Final Itinerary Generation, breakable]
\begin{minted}{markdown}
You are a highly creative and knowledgeable tour guide, specifically to design a perfect day trip itinerary.
Please consider carefully and use the provided "Candidate POIs" list to craft a one-day itinerary in the form of an engaging and realistic travel story.

## Itinerary Information

Next, please follow the guidelines I provide to design a memorable day trip itinerary for tourists.

Design a day trip itinerary for tourists:
- **Order of candidate POIs**: {context_string}
- **Must-see POIs**: {keyword_reqs}
- **Keyword Requirements**: {keyword_reqs}
- **User's Original Request**: {userReqList}
- **Start POI**: {start_poi}
- **End POI**: {start_poi}

## Constraints

- **Itinerary time**: Less than {hours} hours
- **POI selection**: Must follow the given sequence order

## Output Format:
{{
    "itinerary": "List of POIs, separated by '->'"
    "Overall reason": "Overall recommendation reason for the designed day trip itinerary",
    "pois": {{
        "n": "Description and recommendation reason for each POI", ...
    }}
}}

Note:

- "n" is the sequence number, which should be an integer. Sequence numbers in the output must be in ascending order and match the sequence number of the selected POIs from the candidate list.
- "itinerary" lists all the POIs' names visited, separated by '->', such as "poi1->poi2->...", note that it includes names only, without sequence numbers, and the order is consistent with the order of POIs in "pois".


## Pre-action Considerations
1. Work on problems step-by-step.
2. Do not omit or simplify anything.
3. Ensure the tourists feel that the itinerary is tailor-made for them.
4. **ONLY CHOOSE** POIs from the **candidate POIs** list, in ascending order of the **candidate POIs sequence**.
5. The number of cafes and bars cannot exceed two, and they must comply with the sequence order of POIs. **Bars should be placed at the end of the itinerary, and cafes should not be the last stop**.
6. Ensure that every **keyword requirement** is strictly met, for example, if a user mentioned wanting to visit 3 spots, your planned itinerary should strictly include only 3 POIs as per the user's request.


## Itinerary Creation Steps
1. Based on the **candidate POIs** list, select suitable POIs in ascending order to include in the itinerary. Ensure your selection is filtered to choose POIs that compose an itinerary of {hours} hours, not all POIs from the **Order of candidate POIs** list should be included.
2. All included POIs must follow the ascending sequence order of the **Order of candidate POIs**.
3. If the inclusion of a café or bar disrupts the sequence order of POIs, **exclude it from the itinerary**.
4. Ensure every **keyword requirement: {keyword_reqs}** is met by at least one POI in the **candidate POIs sequence**.
5. **User's original requirements: {userReqList}** also need to be seriously considered and ideally met by at least one POI in the **candidate POIs sequence**.
6. Finally, generate a JSON file containing all selected POIs.

Now, following the **Itinerary Creation Steps** and **Itinerary Restrictions**, plan a {hours} hours itinerary.
\end{minted}
\end{tcolorbox}

% \vspace{20pt}
\clearpage

\subsection{Prompt for Baseline}\label{supple:prompt_baseline}

We provide the prompt for the baseline GPT-4 CoT below. We remove the ``Think step by step'' part and ``thoughts'' for baselines without CoT in the output format.

\begin{tcolorbox}[title=Prompt for baseline GPT-4 CoT, breakable]
\begin{minted}{markdown}
You are a travel planning assistant. Please help me plan a city tour itinerary in {city} based on requirements. The output should include specific Points of Interests (POI), and the itinerary should contain no less than 6 POIs:

## User Request
{user_request}

Think step by step: You need to understand and analyze the user's request, including must-see POIs, positive requests, negative requests, etc., and then provide your recommended POIs and reasons for recommendation.

## Output Format:
{{
    "thoughts": "Your understanding and analysis of user requirements",
    "itinerary": "List of POIs, separated by '->'",
    "overall_reason": "The overall reason for recommending this one-day tour itinerary",
    "pois": {{
        "n": "Description and reason for recommending the POI", ...
    }}
}}
n starts from 1 and increments. Please strictly follow the output format to return JSON.
\end{minted}
\end{tcolorbox}

\subsection{Prompt for LLM-evaluated Metrics}

Prompt for LLM-evaluated metrics is provided below.

\begin{tcolorbox}[title=Prompt for LLM-evaluated Metrics, breakable]
\begin{minted}{markdown}
You are a professional travel assistant. I will provide my request for a one-day travel itinerary, and several candidate itineraries containing a list of POIs and descriptions. You should help me compare the itineraries and rank them based on several criteria.

## Criteria
1. POI Quality: how interesting and diverse the POIs are
2. Itinerary Quality: the overall quality and coherence of the itinerary
3. Matchness: the matchness between the itinerary and the user query

## Input Format

Each input candidate itinerary is a dictionary in the following format:

{{
    "itinerary": "a list of POIs, separated by '->'"
    "overall_reason": "The overall recommendation reason for the designed one-day travel itinerary",
    "pois": {{
        "n": "description of the POI", ...
    }}
}}

## Request

{user_request}

## Candidate Itineraries

{itineraries}

## Output Format

Output a json object (dictionary) with four keys: "poi_quality", "itinerary_quality", "matchness", "language_quality". Each value is a list of indexes, representing the rank of the candidates with the corresponding key serves as the criterion (in descending order, i.e. from high to low). For example, '"poi_quality": [4,1,3,2]' suggests that Candidate 4 has the highest POI quality, then Candidate 1 and 3, and Candidate 2 has the lowest POI quality.

Ensure that your output can be parsed with Python json.loads. Do not output anything else.
\end{minted}
\end{tcolorbox}

\end{document}